\documentclass{article}

\usepackage{iclr2023_conference,times}

\usepackage[utf8]{inputenc} 
\usepackage[T1]{fontenc}    
\usepackage{hyperref}       
\usepackage{url}            
\usepackage{booktabs}       
\usepackage{amsfonts}       
\usepackage{nicefrac}       
\usepackage{microtype}      
\usepackage{xcolor}         
\usepackage{multirow}

\title{Neuroevolution is a Competitive Alternative \\ to Reinforcement Learning for Skill Discovery}

\author{
Felix Chalumeau\thanks{Equal Contribution}~~\boldmath{$^1$}~~~~ Raphael Boige\footnotemark[1]~~$^1$~~~~ Bryan Lim$^2$~~~~ Valentin Mac\'e$^1$ \\ \textbf{Maxime Allard}\boldmath{$^2$}~~~~ \textbf{Arthur Flajolet}\boldmath{$^1$}~~~~  \textbf{Antoine Cully}\thanks{Equal Supervision}~~\boldmath{$^2$}~~~~ \textbf{Thomas Pierrot}\footnotemark[2]~~\boldmath{$^1$} \\
 $^1$InstaDeep  $^2$Imperial College\\
 \footnotesize{\texttt{\{f.chalumeau, r.boige, v.mace, a.flajolet, t.pierrot\}@instadeep.com}} \\
 \footnotesize{\texttt{\{bwl116, m.allard20, a.cully\}@ic.ac.uk}} \\
}

\usepackage{amsmath}
\usepackage[ruled]{algorithm2e}
\usepackage{algpseudocode}
\usepackage{graphicx}
\usepackage{wrapfig}
\usepackage{caption}
\usepackage{subcaption}

\usepackage{xspace}
\usepackage{units}

\newcommand{\mymath}[1]{\ensuremath{#1}\xspace}
\def\bbbe{{\rm I\!E}} 
\newcommand{\Esp}{{\bbbe}{}}
\newcommand{\reals}{\mymath{\mathbb R}}

\definecolor{myred}{rgb}{0.8,0,0}
\definecolor{mygreen}{rgb}{0,0.6,0}
\definecolor{myblue}{rgb}{0,0,0.7}

\newcommand{\skilldescriptor}{\phi}
\newcommand{\descriptorspace}{\Phi}

\newcommand{\ns}{{\sc ns}\xspace}
\newcommand{\nslc}{{\sc nslc}\xspace}

\newcommand{\ant}{{\sc ant}\xspace} 
\newcommand{\hc}{{\sc halfcheetah}\xspace} 
\newcommand{\walker}{{\sc walker}\xspace} 
\newcommand{\omnidir}{{\sc omnidirectional}\xspace} 
\newcommand{\unidir}{{\sc unidirectional}\xspace} 
\newcommand{\antmaze}{{\sc ant-maze}\xspace} 
\newcommand{\hcuni}{{\sc halfcheetah-uni}\xspace} 
\newcommand{\antuni}{{\sc ant-uni}\xspace} 
\newcommand{\walkeruni}{{\sc walker-uni}\xspace} 
\newcommand{\antomni}{{\sc ant-omni}\xspace}
\newcommand{\ptmaze}{{\sc point-maze}\xspace}
 
\newcommand{\anttrap}{{\sc ant-trap}\xspace}

\newcommand{\hchurdles}{{\sc halfcheetah-hurdles}\xspace}

\newcommand{\qdpg}{{\sc qd-pg}\xspace}

\newcommand{\tddd}{{\sc td3}\xspace} 
\newcommand{\sac}{{\sc sac}\xspace}

\newcommand{\mees}{{\sc me-es}\xspace}

\newcommand{\qdes}{{\sc qd-es}\xspace}

\newcommand{\ppo}{{\sc ppo}\xspace} 

\newcommand{\nsres}{{\sc nsr-es}\xspace} 
\newcommand{\nsraes}{{\sc nsra-es}\xspace}

\newcommand{\me}{{\sc map-elites}\xspace}
\newcommand{\aurora}{{\sc aurora}\xspace}
\newcommand{\ruda}{{\sc ruda}\xspace}
\newcommand{\pgame}{{\sc pga-map-elites}\xspace}
\newcommand{\pgaaurora}{{\sc pga-aurora}\xspace}
\newcommand{\cmame}{{\sc cma-me}\xspace}
\newcommand{\diayn}{{\sc diayn}\xspace}
\newcommand{\dads}{{\sc dads}\xspace}
\newcommand{\smerl}{{\sc smerl}\xspace}
\newcommand{\diaynsmerl}{{\sc smerl(diayn)}\xspace}
\newcommand{\dadssmerl}{{\sc smerl(dads)}\xspace}
\newcommand{\diaynsum}{{\sc diayn+reward}\xspace}
\newcommand{\dadssum}{{\sc dads+reward}\xspace}

\newcommand{\qd}{{\sc qd}\xspace}
\newcommand{\rl}{{\sc rl}\xspace}
\newcommand{\pgr}{{\sc pg}\xspace}
\newcommand{\mi}{{\sc mi}\xspace}
\newcommand{\mirl}{{\sc mi rl}\xspace}
\newcommand{\es}{{\sc es}\xspace}

\newcommand{\vae}{{\sc vae}\xspace}
\newcommand{\jax}{{\sc jax}\xspace}
\newcommand{\brax}{{\sc brax}\xspace}
\newcommand{\qdax}{{\sc qdax}\xspace}

\iclrfinalcopy
\begin{document}

\maketitle

\begin{abstract}
  
Deep Reinforcement Learning (\rl) has emerged as a powerful paradigm for training neural policies to solve complex control tasks. However, these policies tend to be overfit to the exact specifications of the task and environment they were trained on, and thus do not perform well when conditions deviate slightly or when composed hierarchically to solve even more complex tasks. Recent work has shown that training a mixture of policies, as opposed to a single one, that are driven to explore different regions of the state-action space can address this shortcoming by generating a diverse set of behaviors, referred to as skills, that can be collectively used to great effect in adaptation tasks or for hierarchical planning. This is typically realized by including a diversity term - often derived from information theory - in the objective function optimized by \rl. However these approaches often require careful hyperparameter tuning to be effective. In this work, we demonstrate that less widely-used neuroevolution methods, specifically Quality Diversity (\qd), are a competitive alternative to information-theory-augmented \rl for skill discovery. Through an extensive empirical evaluation comparing eight state-of-the-art algorithms (four flagship algorithms from each line of work) on the basis of (i) metrics directly evaluating the skills' diversity, (ii) the skills' performance on adaptation tasks, and (iii) the skills' performance when used as primitives for hierarchical planning; \qd methods are found to provide equal, and sometimes improved, performance whilst being less sensitive to hyperparameters and more scalable. As no single method is found to provide near-optimal performance across all environments, there is a rich scope for further research which we support by proposing future directions and providing optimized open-source implementations.

\end{abstract}

\section{Introduction}

In the past decade, Reinforcement Learning (\rl) has shown great promise at tackling sequential decision making problems in a generic fashion, leading to breakthroughs in many fields such as games~\citep{silver2017mastering}, robotics~\citep{andrychowicz2020learning}, and control in industrial settings~\citep{degrave2022magnetic}. However, neural policies trained with \rl algorithms tend to be over-tuned to the exact specifications of the tasks and environments they were trained on. Even minor disturbances to the environment, to the starting state, or to the task definition can incur a significant loss of performance~\citep{kumar2020one, pinto2017robust}. A standard approach to improve generalization is to introduce more variations during training~\citep{tobin2017domain} but this assumes we can foresee all possibilities, which is not always true in the real world. Even in settings where this is feasible, introducing a wide spectrum of variations will make the problem harder to solve and the resulting policy may not perform as well in the nominal case. Another approach consists in co-training an adversarial agent whose task is to perturb the environments so as to minimize the policy's performance~\citep{pinto2017robust}. However, adversarial methods are notoriously unstable in Deep Learning~\citep{arjovsky2017towards} and can also compromise performance in the nominal scenario.

To improve robustness without explicitly identifying all possible variations, jointly training multiple policies to solve the same task in diverse ways has emerged as a promising line of work in the \rl literature~\citep{kumar2020one}. To motivate the approach, consider the problem of learning a policy to control the joints of a legged robot with the goal of running as fast as possible. Any damage to the robot legs might affect an optimal policy's ability to make the robot run fast, if at all. Yet, many of the slightly sub-optimal policies to the original problem (e.g. a policy making the robot hop using only one leg) would perform equally well in this perturbed setting. Two seemingly-opposed main lines of work have been pursued to maximize both performance and diversity in a collection of policies.

\rl-rooted approaches~\citep{eysenbach2018diversity, sharma2019dynamics, kumar2020one} introduce a randomly-generated latent variable and parametrize the policy to be a function of the state as well as this latent variable. At training time, the latent variable is drawn from a static distribution and fed as an input alongside the state to the policy, effectively defining a mixture of policies. To encourage diversity among these policies, a term derived from information theory that depends both on the policy parameters and the latent variable is added to the objective function (hereinafter referred to as fitness function). This term is typically formulated as the mutual information between the latent variable and a subset of the policy's trajectory, possibly conditioned on observations from the past.

Neuroevolution-rooted approaches instead stem from the subfield of Quality Diversity (\qd) optimization \citep{pugh2016quality, cully2017quality, chatzilygeroudis2021quality} and combine the tools developed in this space with \rl algorithms to get the best of both worlds~\citep{nilsson2021policy,pierrot2022diversity}. \qd optimization aims at generating and maintaining large and diverse collections of solutions, as opposed to a single optimal solution in Optimization Theory, by imitating the natural evolution of individuals competing for resources in their respective niches. In comparison to traditional Evolutionary Strategies, \qd algorithms explicitly use a mapping from solution to a vector space, referred to as behavior descriptor space, to characterize solutions and maintain a data structure, a \emph{repertoire}, filled with high-performing solutions that cover this space as much as possible.

Evolutionary Strategies (possibly hybridized with \rl algorithms) have proven to be a competitive alternative to \rl algorithms for many common sequential-decision making problems~\citep{pierrot2022diversity, salimans2017evolution}. Hence, it is natural to believe that \qd algorithms could also be competitive with information-theory-augmented \rl approaches to generate diverse populations of high-performing policies in similar settings. Yet, \qd approaches remain neglected in skill-discovery studies~\citep{kumar2020one}, perhaps because they lack the sample-efficiency of state-of-the-art \rl algorithms, sometimes requiring two orders of magnitude more interactions with the environment to solve a task~\citep{pierrot2022diversity}. While this is a significant shortcoming for real-world applications that cannot be accurately described by a computational model, simulators are readily available for many applications. Additionally, when the simulator and the algorithm are implemented using modern vectorized frameworks such as \jax~\citep{jax2018github} and \brax~\citep{freeman2021brax}, evolutionary approaches are competitive with \rl approaches in terms of total training time on an accelerator in spite of the low sample-efficiency of these methods~\citep{lim2022accelerated}.

Our contributions are the following. \textbf{(1.)} We provide extensive experimental evidence that \qd methods are competitive with \rl ones for skill discovery in terms of performance given fixed compute and training time budgets and hyperparameter sensitivity. Specifically, using environments taken from the \qd and \rl literature, we compare eight state-of-the-art skill-discovery methods from the \rl and \qd world on the basis of (i) metrics directly evaluating the skills' diversity, (ii) the skills' performance on adaptation tasks, and (iii) the skills' performance when used as primitives for hierarchical planning. \textbf{(2.)} We open source efficient implementations of all environments and algorithms\footnote{ \footnotesize{\url{https://github.com/instadeepai/qd-skill-discovery-benchmark}}} based on the QDax library\footnote{\vspace{-0.5cm}\url{https://github.com/adaptive-intelligent-robotics/QDax}}. Armed with these, running any of the experiments, some of which require hundreds of millions of environments steps, takes only 2 hours on a single affordable accelerator. \textbf{(3.)} We provide a detailed analysis of the strengths and weaknesses of all methods, we show that no single method outperforms all others on all environments, and we identify future research directions.

\begin{figure}
    \centering
    \includegraphics[width=1.0\textwidth]{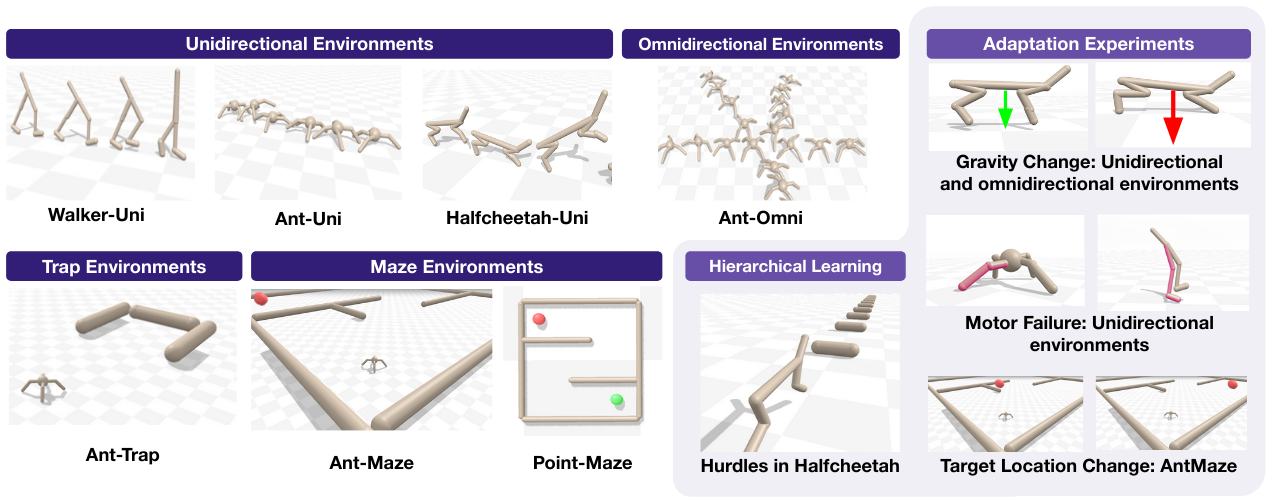}
    \caption{Illustrations of environments used for benchmarking (left) and perturbations applied to them for adaptation and hierarchical learning experiments (right). In \unidir environments, policies are trained to make the robots run forward with diverse gaits. In the \omnidir environment, the goal is to make a legged robot move on the 2D plane while minimizing the control energy. In \anttrap, a robot learns to run in the forward direction while avoiding a trap. In \antmaze and \ptmaze, a robot must reach a target position in a maze. Once trained over these tasks, policies are evaluated  on perturbed versions of the tasks and environments.}
    \label{fig:bench_summary}
    \vspace*{-0.5cm}
\end{figure}

\section{Preliminaries and Problem Statement}
\label{sec:preliminaries}
We consider sequential decision making problems formulated as Markov Decision Processes (MDPs) and defined by $(\mathcal{S}, \mathcal{A}, \mathcal{R}, \mathcal{P}, \gamma)$,  where $\mathcal{S}$ is the state space, $\mathcal{A}$ is the action space, $\gamma \in [0,1]$ is the discount factor, $\mathcal{R}: \mathcal{S} \times \mathcal{A} \rightarrow \reals$ is the reward signal and $\mathcal{P}: \mathcal{S} \times \mathcal{A} \rightarrow \mathcal{S}$ is the transition function. Policies are assumed to be implemented by neural networks. We consider two cases. \textbf{(1.)} \textit{A mixture of policies is conditioned on a latent variable} $z \in \mathcal{Z}$, in which case it is denoted by $\pi_{\theta}: \mathcal{S} \times \mathcal{Z} \rightarrow \mathcal{D}(\mathcal{A})$ where $\theta$ denotes the neural network weights and $\mathcal{D}(\mathcal{A})$ is the space of distributions over the action space. The collection of policies $\Pi$ is obtained by enumerating all possible values for the latent variable. \textbf{(2.)} \textit{Policies share the same neural network architecture but have their own sets of parameters.} In this case, $\Pi$ corresponds to the collection of $N$ policies $\pi_{\theta_i}: \mathcal{S} \rightarrow \mathcal{D}(\mathcal{A}), \ i \in [1,N]$. We denote the trajectory of a single policy $\pi$ in the environment by $\tau \in \Omega$. The expected sum of rewards of a policy $\pi$, referred to as its fitness, is $F(\pi) = \Esp_{\tau}{\sum_t \gamma^t \mathcal{R}(s_t, a_t)}$.

In this work, compared to traditional \rl, we consider a second objective: the policies in the collection $\Pi$, referred to as \emph{skills}, must not only be high-performing but also diverse as a whole. The notion of diversity can be defined in various ways. In the \rl literature, diversity is often measured by looking at the volume spanned by components of the trajectories induced by the policies, such as the sequences of visited states~\citep{parker2020effective}. In the \qd literature, diversity is measured in a so-called \emph{behavior descriptor space} $\descriptorspace$ using a mapping $\skilldescriptor: \Omega \rightarrow \descriptorspace$ that is either hand-engineered or progressively refined in an unsupervised fashion. This approach is more general as it includes as a special case the first one but also allows to leverage expert knowledge on the task at hand, which is often available. We use the notion of behavior descriptor space in order to quantitatively compare the diversity of policies generated by \rl and \qd methods. Specifically, we define a bounded behavior descriptor space for each MDP, that may differ from the ones used internally by \qd methods, which is discretized into a discrete tessellation of cells. Policies from $\Pi$ are inserted into the tessellation as a function of their behavior descriptors which enables us to use diversity metrics developed in the \qd literature. This process is made precise in Section \ref{sec:diversity}.

\section{Different solutions to the same problem}
\label{sec:diversity-methods}

\textbf{Information-theory-augmented RL methods.} Most methods aiming to maximize diversity in the \rl literature, such as \diayn~\citep{eysenbach2018diversity} and \dads~\citep{sharma2019dynamics}, fall into the class of latent-conditioned policies introduced in Section \ref{sec:preliminaries}. They strive to maximize the Mutual Information (\mi) between a latent variable $z$, which is drawn randomly from a pre-specified distribution, and some components of the trajectories induced by the conditioned policy. The \mi maximization is approximately carried out through the introduction of a discriminator function $z \mapsto q(z|s)$ given a state $s \in \mathcal{S}$ (resp. a dynamics function $z \mapsto q(s_{t+1}|s_{t}, z)$ given the current state $s_{t}$ and the next state $s_{t+1}$) for \diayn (resp. \dads) trained separately to differentiate the latent variables that lead to visit different states. The latent-conditioned policy is optimized using an off-policy model-free \rl algorithm by taking the reward as the discriminator value, referred to as \emph{intrinsic reward}.

\diayn and \dads are unsupervised algorithms as they do not aim to maximize a reward function associated with a specific task. Recent work in \mi-based \rl, namely \smerl~\citep{kumar2020one}, has extended \diayn and \dads to supervised settings where both diversity and performance w.r.t. a reward function should be optimized. They formulate the problem as a Constrained MDP (CMDP) where the objective is to maximize the diversity of a mixture of policies, while also constraining the learned policies to be close to optimality w.r.t. the reward function. By solving these CMDPs, they obtain a collection of policies that are both diverse and high-performing. \smerl can use either \diayn or \dads internally to generate collections of policies, two variants we refer to as \diaynsmerl and \dadssmerl. An alternative approach to \smerl simply consists in summing the task reward and the intrinsic reward and using the original \diayn and \dads algorithms~\citep{kumar2020one, osa2021discovering, gaya2021learning}. This approach is very similar to \citet{hausman2018learning}. We refer to these variants as \diaynsum and \dadssum. For clarity's sake, we only report the performance of the original \diayn (resp. \dads) algorithm in the Appendix as it turns out to be significantly outperformed by \diaynsum (resp. \dadssum) in our experiments.

\textbf{QD methods} fall into the second category of policies introduced in Section \ref{sec:preliminaries}. These methods explicitly evolve and maintain a discrete repertoire of independent policies. \qd algorithms traditionally rely on Evolutionary algorithms to incrementally update policies almost independently. Such techniques have the advantage of being easily parallelizable and can be used with non-differentiable policies and controllers~\citep{tang2020neuroevolution}. They have been shown to be viable alternatives to \rl algorithms~\citep{salimans2017evolution}.

\me is one of the most widespread \qd algorithms. This method assumes that the behavior descriptor space has been discretized into a finite tessellation of cells, which together define a \emph{repertoire} where policies are stored. This algorithm proceeds by iterations. At each iteration, a batch of policies is sampled uniformly from the repertoire and copied. Mutations and crossover operations are applied to them to obtain a new batch of policies which are evaluated in the environment to compute their fitnesses and behavior descriptors. For any of these policies, if the cell in the repertoire corresponding to its behavior descriptor is empty, then it is copied into this cell. Otherwise, the new policy replaces the current incumbent only if its fitness is higher than the current incumbent's fitness. 

The exact choice of mutation and crossover operators significantly affects the performance of \qd algorithms. Choices that rely exclusively on random perturbations typically require a significant number of iterations. To remedy this problem, several works have combined first-order optimization methods with Evolutionary algorithms to guide the parameter search~\citep{colas2020scaling, pierrot2022diversity}. In particular, \pgame \citep{nilsson2021policy} is a hybrid algorithm that builds on the \me framework to select, mutate, and replace policies in the repertoire but further introduces a policy-gradient mutation operator which is used alongside the genetic operator. As part of every \me iteration, \pgame collects the experience of all evaluated policies and stores it into a replay buffer in order to train a pair of critics using the \tddd algorithm~\citep{fujimoto2018addressing}. These critics are used to compute the policy-gradient updates applied to the policies.

While \me and \pgame are provided with a pre-specified behavior descriptor space, it is defined in an unsupervised fashion in a recent extension to \me dubbed \aurora~\citep{cully2019aurora, grillotti2022aurora}. This is achieved by training an autoencoder to reconstruct the trajectories collected in the environments and defining the behavior space as the latent space of the autoencoder. In this paper, we introduce a new unsupervised method which extends both \aurora and \pgame. This method, referred to as \pgaaurora, use the same policy-gradient mutation operator as \pgame and the same unsupervised technique as \aurora to define the behavior descriptor space.

\section{Related Work}

Historically, \qd algorithms originated from Evolutionary algorithms, which also implement selection, mutation, and evaluation of populations of solutions. Novelty Search (\ns) \citep{lehman2011abandoning} first introduced the idea of searching for novelty alone. In a sense, \ns is similar to \diayn and \dads as it is an exploration-only algorithm. Novelty Search with Local Competition (\nslc) \citep{lehman2011evolving} and \me \citep{mouret2015illuminating} are two of the most popular \qd algorithms. \nslc builds off the \ns algorithm and, in contrast to \me, it maintains an unstructured repertoire of solutions that are selected for their performance when compared to similar individuals. 
Relying on the same diversity-seeking strategies, \qdes algorithms, which combine \qd and Evolutionary Strategies (\es), such as \nsres and \nsraes, have been successfully applied to challenging continuous control environments~\citep{conti2018improving}. Aiming for improved sample efficiency, \citet{colas2020scaling} developed \mees which optimizes both for quality and diversity using \me and two \es populations. Similarly, \cmame~\citep{fontaine2020covariance} uses a covariance matrix adaptation technique to model and update a population distribution to sample individuals from and also maintains a population of varied emitters to generate solutions. More recently, some authors proposed to incorporate a policy-gradient component in the evolutionary framework~\citep{nilsson2021policy, pierrot2022diversity, tjanaka2022approxdqd}. Inspired by \pgame, the authors of \qdpg~\citep{pierrot2022diversity} introduced a diversity policy gradient to make diversity information available at the timestep level.

Multiple methods from the \rl literature build an intrinsic reward with \mi to learn a diverse set of skills. They are similar in spirit to \diayn and can be adapted, along the same lines as what is done for \smerl, to use extrinsic rewards. \citet{Gregor2016} originally introduced \mi for skill discovery. Most recent efforts in the skill-discovery \rl literature focus on ways to extend \diayn and \dads to improve the state coverage of the learned skills by using contrastive learning~\citep{laskin2022cic}, adding constraints to the skills~\citep{hansen2021entropic, kamienny2021direct}, or changing the loss function~\citep{durugkar2021wasserstein}. Nevertheless, \citet{zahavy2021discovering} introduced an original approach where the intrinsic rewards are defined based on successors features rather than \mi. We choose \diayn, \dads, and \smerl as baselines in our experiments as they are the most widely used \rl methods for skill discovery at the time of writing. There also exist methods based on tools other than \mi that strive to evolve a diverse population of \rl agents, such as \citet{zhou2022continuously, parker2020effective, zhang2019learning, masood2019diversity}, but these studies mostly consider diversity as a means to increase performance w.r.t. the rewards, whereas \diayn, \smerl, and \dads are evaluated on pure diversity, on adaptation tasks, or in hierarchical learning settings.

\vspace*{-0.1cm}
\section{Experiments}
\label{sec: exp}
In this section, we present the frameworks used to compare the methods listed in Section \ref{sec:diversity-methods} and analyze the results. First, in Section \ref{sec:domains}, we introduce the environments and tasks upon which the frameworks are built. In Section \ref{sec:diversity}, we leverage tools and metrics from the \qd literature to directly evaluate the diversity of the solutions found by each method in the process of solving the tasks. In Section \ref{sec:downstream}, we investigate how diversity translates into robustness with few-shot adaptation experiments inspired from the \rl literature. Finally, in Section \ref{sec:hierarchical}, we evaluate the policies found by each method when used as primitives in a hierarchical learning setting.

\subsection{Environments}
\label{sec:domains}

In order not to favor any method in the study, we use three different categories of environments and tasks: (i) continuous control locomotion tasks from the \rl literature, (ii) exploration tasks with deceptive rewards from the \qd literature, and (iii) hybrid tasks mixing both types as a middle ground. All environments are illustrated on Figure \ref{fig:bench_summary}.
 
\textbf{Low-dimensional exploration task with deceptive rewards.} Following several works from both the skill-discovery \rl and \qd communities~\citep{eysenbach2018diversity, kamienny2021direct, campos2020edl}, we consider a simple navigation environment where a point is controlled to navigate on a 2D plane to a specified target position. To turn the task into a challenging exploration problem, we introduce a maze through which the point must navigate to reach the target position. We refer to this task as \ptmaze.

\textbf{High-dimensional locomotion tasks.} We consider three locomotion tasks, namely \ant, \hc and \walker, often used in the \rl community to evaluate algorithms, including in \rl works focused on diversity-seeking strategies~\citep{nilsson2021policy, eysenbach2018diversity, fujimoto2018addressing}. In these environments, the problem is to move legged robots by applying torques to their joints through actuators. These tasks are challenging for evolutionary algorithms: they typically require orders of magnitude more environment interactions than \rl methods to evolve competitive policies~\citep{colas2020scaling, pierrot2022diversity}. We define two subsets of tasks for these environments. The first subset, referred to as \unidir locomotion tasks and with the reward signal defined as the x-axis velocity minus the energy consumption, includes \antuni, \walkeruni, and \hcuni. The second subset, referred to as \omnidir locomotion tasks and with the reward signal defined as the opposite of the energy consumption, includes \antomni.
 
\textbf{Mixing locomotion and exploration tasks.} Following \citet{pierrot2022diversity, parker2020effective, frans2017meta, shi2020efficient}, we consider two additional hybrid tasks built upon \ant that exhibit both continuous control and exploration challenges, making them particularly difficult even for state-of-the-art evolutionary and \rl algorithms. In \anttrap, the ant has to run forward while initially facing a trap consisted of three walls that prevents it from moving forward along the x-axis direction indefinitely. In \antmaze, the ant is initially placed in a maze and the goal is to reach a specified target position.

\subsection{Experiments directly evaluating diversity}
\label{sec:diversity}

\begin{figure}
    \centering
    \includegraphics[width=\textwidth]{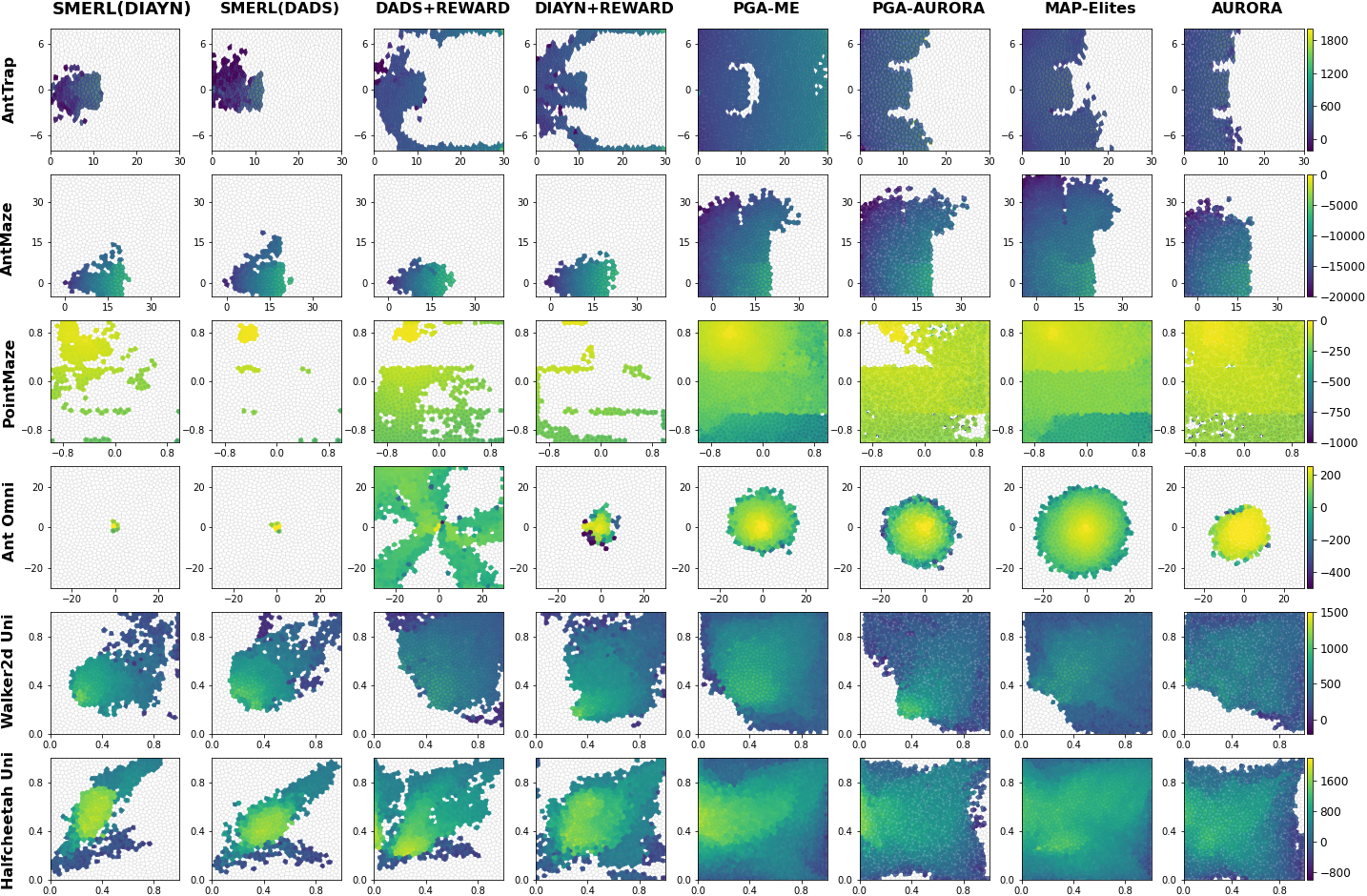}
    \caption{Visualizations of repertoires post training. Each repertoire divides the behavior descriptor space into 1024 cells. Cells are left blank when no policy with the corresponding behavior descriptor has been found and are otherwise colored as a function of the best fitness achieved across policies with a matching behavior descriptor. For \anttrap, \antmaze, \ptmaze, and \antomni, the behavior descriptor is the $(x, y)$ position reached at the end of an episode: these plots highlight explored areas of the 2D plane. For \hcuni and \walkeruni, the descriptor is the fraction of time the robots' feet were touching the ground during an episode (one dimension per foot).}
    \label{fig:repertoires}
    \vspace*{-0.5cm}
\end{figure}

On the one hand, metrics quantifying the amount of diversity in a collection of policies are lacking in the \mirl literature. Success in inducing diversity is instead typically assessed indirectly through performance on adaptation or hierarchical learning experiments where either the tasks are modified or new ones are introduced \citep{eysenbach2018diversity, kumar2020one, sharma2019dynamics}. On the other hand, diversity metrics were introduced very early in the \qd literature and have been refined over time since \qd algorithms are primarily assessed on this basis. Hence, we use metrics developed in the \qd literature and extend them to \mirl approaches to be able to compare the methods on the same ground. Noting that almost all \qd metrics crucially rely on the definition of a behavior descriptor space (to characterize solutions) and the availability of a repertoire of solutions indexed with respect to this space for all methods, we specify behavior descriptors for all tasks and extend the concept of repertoires to \mirl methods.

\textbf{Behavior descriptor spaces.} For \ptmaze, \anttrap, \antmaze, and \antomni, following several prior \qd and \rl works \citep{pierrot2022diversity, hansen2021entropic}, the behavior descriptor of a policy is defined as the $(x, y)$ position of the robot's center of gravity at the end of an evaluation episode. For \unidir tasks, the $i$-th component of the behavior descriptor vector is defined as the proportion of time during which the $i$-th foot of the robot is in contact with the ground in an evaluation episode. This is a simple but effective way to capture the gait of a robot as demonstrated in \citet{cully2015robots} where it is used to help robots adapt after having some of their legs damaged. 

We stress that, in our experiments, \me and \pgame make use of the same behavior descriptors used to compute the \qd metrics, which induces a bias in the analysis. This bias does not affect \aurora and \pgaaurora as these methods build their own behavior descriptors in an unsupervised fashion. Nevertheless, for fairness, we embed the choice of the behavior descriptor space into \mirl methods whenever possible. Hence, in \antmaze, \anttrap, \antomni and \ptmaze, \mirl algorithms are provided with the prior that diversity w.r.t. the $(x, y)$ position of the robot is what needs to be optimized. Implementation details are deferred to the Appendix. 

\begin{table}
  \caption{Maximum fitness and \qd score at the end of the training phase (median over 5 seeds). Additional statistics (mean, standard deviation, and interquartile mean) are included in the Appendix.}
  \centering
  \resizebox{.98\textwidth}{!}{
  \label{qd-score-comparison}
  \centering

  \begin{tabular}{lccccccc}
    \toprule
         & \anttrap & \antmaze & \ptmaze & \antuni & \antomni & \walkeruni & \hcuni \\
    \midrule
        \multicolumn{8}{c}{\textbf{Maximum fitness}} \\
    \midrule
    \diaynsmerl & 4.49 \text{e}2 & -7.88 \text{e}3 & -4.00 \text{e}1 & \textbf{1.71 e3} & \textbf{2.49 e2} & 1.19 \text{e}3 & 1.89 \text{e}3 \\
    \dadssmerl & 4.48 \text{e}2 & \textbf{-7.62 e3} & -8.10 \text{e}1 & 1.61 \text{e}3 & \textbf{2.49 e2} & 1.17\text{e}3 & 1.92 \text{e}3 \\
    \dadssum & \textbf{1.28 e3} & -7.81 \text{e}3 & -8.10 \text{e}1 & 1.24 \text{e}3 & 2.38 \text{e}2 & 6.30 e2 & 1.89 \text{e}3 \\
    \diaynsum & 9.42 \text{e}2 & -8.13 \text{e}3 & -4.50 \text{e}1 & 1.34 \text{e}3 & 2.29 \text{e}2 & 1.00 \text{e}3 & \textbf{2.13 e3} \\
    \pgame & 9.77 \text{e}2 & -9.32 \text{e}3 & \textbf{-2.30 e1}  & 1.45 \text{e}3 & \textbf{2.49 e2} & \textbf{1.22 e3} & 1.75 e3 \\
    \pgaaurora & 4.72 \text{e}2 & -9.27 \text{e}3 & -6.3 \text{e}1 & 1.44 \text{e}3  & \textbf{2.49 e2} & 1.05 \text{e}3 & 1.47 \text{e}3 \\
    \me & 3.81 \text{e}2 & -1.01 \text{e}4 & -2.5 \text{e}1 & 4.25 \text{e}2  & \textbf{2.49 e2} & 6.87 \text{e}2 & 1.23 \text{e}3 \\
    \aurora & 4.05 \text{e}2 & -1.10 \text{e}4 & -2.35 \text{e}1 & 5.05 \text{e}2  & \textbf{2.49 e2} & 5.71 \text{e}2 & 1.21 \text{e}3 \\
    \midrule
        \multicolumn{8}{c}{\textbf{QD score}} \\
    \midrule
    \diaynsmerl & 1.02 \text{e}5 & 7.84 \text{e}5 & 38.9 \text{e}4 & 6.20 \text{e}4 & 1.95 \text{e}3 & 3.01 \text{e}5 & 7.92 \text{e}5 \\
    \dadssmerl & 9.88 \text{e}4 & 8.22 \text{e}5 & 3.57 \text{e}3 & 0.83 \text{e}5  & 4.15 \text{e}3  & 2.69 \text{e}5& 7.51 \text{e}5 \\
    \dadssum & 2.42 \text{e}5 & 9.12 \text{e}5 & 0.42 \text{e}4 & 2.88 \text{e}5 & 3.05 \text{e}5 & 3.89 \text{e}5 & 8.31 \text{e}5\\
    \diaynsum & 2.27 \text{e}5 & 9.18 \text{e}5 & 4.04 \text{e}4 & 1.29 \text{e}5 & 4.00 \text{e}4 & 4.92 \text{e}5 & 1.68 \text{e}6\\
    \pgame & \textbf{7.89 e5} & 2.74 \text{e}6 & 3.92 \text{e}5 & \textbf{9.19 \text{e}5} & 1.58 \text{e}5 & \textbf{8.23 e5} & 2.98 \text{e}6 \\
    \pgaaurora & 4.40 \text{e}5 & 2.59 \text{e}6 & 2.88 \text{e}5 & 7.33 \text{e}5 & 1.69 \text{e}5 & 5.10 \text{e}5 & 2.39 \text{e}6 \\
    \me & 4.53 \text{e}5 & \textbf{2.98 \text{e}6} & 4.22 \text{e}5 & 9.08 \text{e}5 & 3.04 \text{e}5 & 6.45 \text{e}5 & \textbf{3.05 \text{e}6} \\
    \aurora & 3.54 \text{e}5 & 2.18 \text{e}6 & \textbf{4.58 \text{e}5} & 5.82 \text{e}5 & \textbf{4.00 \text{e}5} & 4.59 \text{e}5 & 2.45 \text{e}6 \\
    \bottomrule
  \end{tabular}
}
  \label{tab:max_fitness}
  \vspace*{-0.4cm}
\end{table}

\textbf{Passive repertoires for RL methods.} The \mirl methods considered in this paper do not maintain a repertoire to store diverse solutions. For fairness, and since increasing the size of the discrete latent space to match the size of the \qd repertoires makes these methods unstable~\citep{eysenbach2018diversity}, we fill a repertoire on the side during training with all the intermediate policies found by these methods, similarly to what is done internally in \qd methods. The repertoires of size 1024 are identically defined across methods using Centroidal Voronoi Tessellations~\citep{vassiliades2017using}. 

\textbf{QD Metrics.} Armed with behavior descriptor spaces for all environments and repertoires for all methods, we compute three metrics that are often used in the \qd community to track the performance of a collection of solutions: (i) the \textbf{maximum fitness} of the repertoire, computed as the maximum fitness attained across the solutions stored in the repertoire, (ii) the \textbf{coverage} of the repertoire, computed as the number of cells (behavior niches) that have been filled, and (iii) the \textbf{QD score}, computed as the sum of fitnesses attained by the solutions stored in the repertoire.

\begin{wrapfigure}{r}{0.45\textwidth}
\vspace*{-0.6cm}
    \centering
    \includegraphics[width=0.45\textwidth]{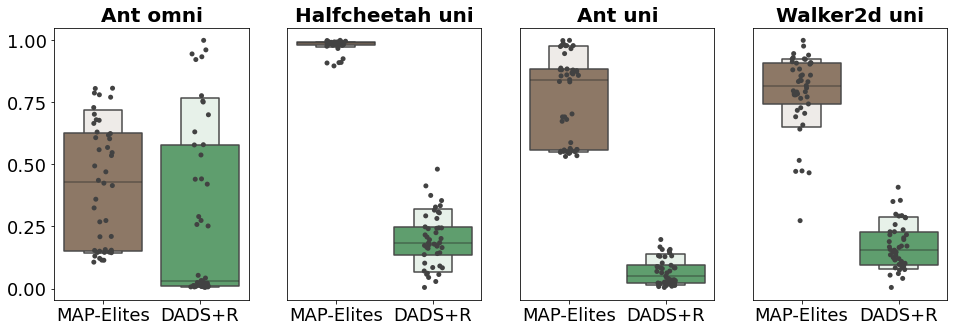}
    \caption{Distribution of \qd scores obtained over a range of hyperparameters on \unidir tasks and \antomni.}
    \label{fig:hyperparam_se}
    \vspace*{-0.5cm}
\end{wrapfigure}

\textbf{Results and analysis.} All methods are trained for exactly two hours on a Quadro RTX 4000 GPU for each environment described in Section \ref{sec:domains}. Each experiment is repeated with 5 seeds and the final repertoire is evaluated using the aforementioned \qd metrics. The maximum fitness and \qd score are reported in Table \ref{tab:max_fitness}. Out of the 5 seeds, the best - in terms of \qd score - final repertoires are depicted on Figure \ref{fig:repertoires} for all methods and environments.

First, observe that, setting aside \antomni where \dadssum has managed to largely explore the environment, \qd methods outperform the \mirl methods for all environments and tasks if we base the comparison on \qd scores and coverage only. This is a particularly desirable property in \antmaze (resp.  \anttrap) because reaching a high coverage of the repertoire means that the method has been able to explore the maze (resp. get past the trap) effectively given that the behavior descriptor vector is defined as the final $(x, y)$ position of the robot.

Second, note that no single method outperforms all others in terms of maximum fitness. \pgame ranks first on \ptmaze and \walkeruni but is always outperformed by an \mirl method in the others. Additional observations are included in the Appendix.

While the excellent performance of \dadssum on \antomni is an impressive achievement in itself, we run an additional set of experiments to investigate whether this may be the result of a lucky guess on the exact choice of hyperparameters. Specifically, we run the same experiments on \antomni and \unidir environments multiple times for \dadssum and \me, sampling from a grid of reasonable hyperparameters values for both methods (exact details are deferred to the Appendix). The medians and quartiles of the \qd-score distributions obtained are shown on Figure \ref{fig:hyperparam_se}. We observe that the performance of \dadssum is very sensitive to the exact choice of hyperparameters and that the median performance of \me is always significantly better than the median performance of \dadssum on all tasks.

\vspace*{-0.15cm}
\subsection{Few-shot adaptation experiments}
\label{sec:downstream}
\vspace*{-0.15cm}

\begin{figure}
    \centering
    \includegraphics[width=1\linewidth]{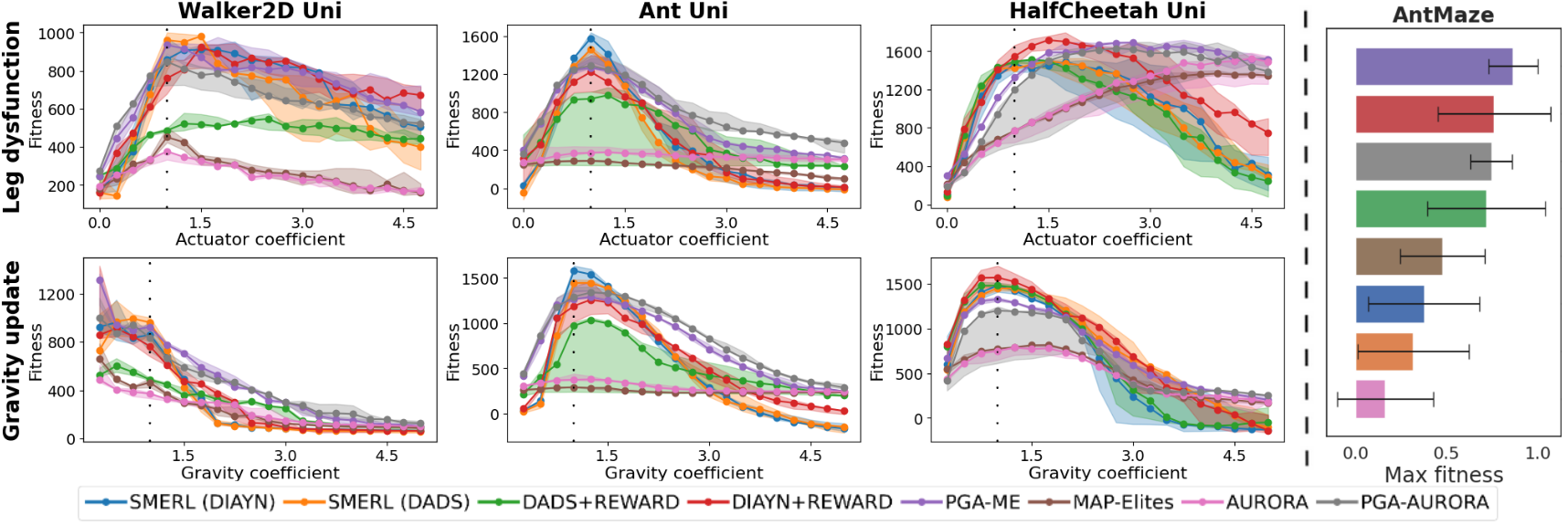}
    \caption{Maximum fitness of the methods when confronted to (left) a perturbed environment and (right) a modified task (median and interquartile range on 5 seeds, additional statistics are in the Appendix). We observe varying resilience levels across methods. In particular, \pgame and \pgaaurora adapt better to extreme environment perturbations in \unidir environments.}
    \label{fig:adapt_exps}
    \vspace*{-0.6cm}
\end{figure}

While metrics borrowed from the \qd literature provide a first basis for comparison, they suffer from an inherent bias in favor of \qd methods as discussed in Section \ref{sec:diversity}. Following prior work that analyze the resilience of \rl methods to changes in the environment or task definition~\citep{kumar2020one}, we design two series of few-shot adaptation experiments that do not favor any particular method a priori. Few-shot here refers to the fact that no re-training is allowed upon modification of the task or the environment, only evaluations of pre-computed policies are allowed. Throughout this section, \me and \pgame use the behavior descriptor spaces introduced in Section \ref{sec:diversity}.

\textbf{Environment adaptation experiments.} In the first series of experiments, we modify a property of the \unidir environments after training, as illustrated on Figure \ref{fig:bench_summary}. First, we vary the gravity coefficient in a neighborhood grid of size 20 around the default value (i.e. 1). We jointly refer to these experiments as the gravity-update setting. Second, we vary the actuators input-to-torque coefficients for a whole leg of the robot in a neighborhood grid of size 20 around the default value (i.e. 1). We jointly refer to these experiments as the leg-dysfunction setting.

\textbf{Task adaptation experiments.} In the second series of experiments, we modify the task of the \antmaze environment by moving the target position after training, which directly modifies the reward signal. New target positions are sampled uniformly on the 2D plane delimited by the maze.

For both series of experiments, we train all methods on the nominal versions of the environments for exactly two hours and re-evaluate the final output of the training phase on the modified environment or task. Specifically, for each method, we enumerate all their learned skills, evaluate them one hundred times in the modified environment, and pick the one that performs best in terms of median fitness across evaluations. Each experiment is repeated with 5 different seeds and the maximal fitnesses obtained are reported on Figure \ref{fig:adapt_exps} for the environment and task adaptation experiments.

\textbf{Results and analysis.} First, note that, on the environment adaptation experiments, \pgame and \pgaaurora are performing on par with \mirl methods in the case of small perturbations, i.e. when the coefficient is perturbed by less than 50\%. Methods that do not leverage policy gradients (i.e. \me and \aurora) are outperformed on all tasks in this regime.

Second, \qd methods are more resilient than \mirl when confronted with significant perturbations. For instance, in the gravity-update setting, the relative performance loss is limited to 40\% for \pgame when the gravity coefficient is divided by 4 compared to at least 60\% for \mirl methods. As a testament of the resilience of \qd methods, note that the best performing method is a \qd one for 10 out of the 12 most extreme scenarios. Similar observations hold for \antmaze where \pgame and \pgaaurora perform as well as \diaynsum and \dadssum.

Third, it is remarkable that \pgaaurora and \aurora perform almost as well as their \qd supervised counterparts in these experiments. This shows that unsupervised behavior descriptor learning is really effective for \qd methods. Furthermore, observe that \pgaaurora adapt with a similar loss in fitness than \mirl methods (recall that all of these methods have access to the exact same priors).

\vspace*{-0.2cm}
\subsection{Hierarchical learning experiments}
\label{sec:hierarchical}
\vspace*{-0.1cm}

To further assess the skills discovered by the methods during training on a nominal environment, we introduce a hierarchical learning setting inspired from the \rl literature ~\citep{eysenbach2018diversity}. In \hchurdles, hurdles are spaced out along the x-axis direction in front of the robot's initial position in \hc, see Figure \ref{fig:bench_summary}. The robot has to jump over them to move forward, the reward definition remaining unchanged. Just like in \citet{eysenbach2018diversity}, we train a \ppo meta-controller~\citep{schulman2017proximal} to choose a skill, which is to be unrolled in the environment for the next ten consecutive timesteps, among the ones learned by a given skill-discovery method during training on the nominal version of the environment, see Section \ref{sec:diversity}. The meta-controller is trained for 200 million policy interactions, corresponding to 20 million actions of the meta-controller.

\begin{wrapfigure}{r}{0.35\textwidth}
\vspace*{-0.6cm}
    \centering
    \includegraphics[width=0.35\textwidth]{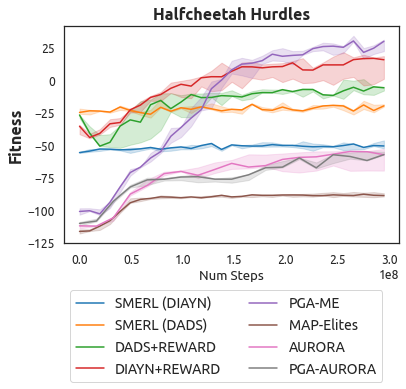}
    \vspace*{-0.5cm}
    \caption{Fitness of the meta-controller as a function of environment steps for all types of skills primitives (median on 5 seeds).}
    \label{fig:hurdles}
    \vspace*{-1.1cm}
\end{wrapfigure}

\textbf{Results and analysis.} The performance of the meta-controller is shown on Figure \ref{fig:hurdles} for various choices of skill-discovery methods used to generate primitives. By visualizing the meta policies in the environment, we observe that \ppo is able to learn a meta-controller that consistently manages to jump over multiple consecutive hurdles when it is trained with primitives derived from \pgame and \diaynsum, which achieve similar performance on this experiment. Surprisingly, all other skill-discovery methods fail to provide primitives that can be composed to make the robot jump over hurdles.

\vspace*{-0.1cm}
\section{Discussion and Conclusion}
\vspace*{-0.1cm}

Our experiments show that \qd methods are competitive with \mirl methods when the goal is to find a set of diverse and high-performing policies in continuous control and exploration environments. \mirl methods are task-agnostic and data-efficient but struggle to generate large collections of policies. Moreover, they are limited in the set of descriptors priors they can accommodate. \qd methods are less data-efficient but can better leverage the vectorization capabilities of modern libraries such as \jax. Moreover, they are significantly less sensitive to the choice of hyperparameters than \mirl methods and their ability to evolve large sets of diverse solutions is a precious tool for hierarchical learning and adaptation applications when the task definition changes or when confronted with extreme environment perturbations. Our benchmarks however show that \qd methods do not outperform \mirl methods and vice-versa, which opens the door to promising work directions. We believe that \mirl methods could make use of skill repertoires to improve their memory capacity thereby increasing the pool of available options for downstream applications. Conversely, \qd methods leveraging gradients, such as \pgame, could benefit from the introduction of intrinsic rewards described in the \mirl literature to guide the search for diversity. Ultimately, we believe that combining the coverage capacity of a repertoire with the data-efficiency of gradient-based techniques is key to develop robust, diverse, and high-performing repertoires of solutions for challenging continuous control.

\section{Acknowledgments}
Research supported with Cloud TPUs from Google's TPU Research Cloud (TRC).

\bibliographystyle{iclr2023_conference}
\bibliography{main}

\appendix

\section{Implementation Details}
In this section we detail the implementation choices that have been made to compare the algorithms. All our implementations are based on the \jax framework \cite{jax2018github} and are specifically designed to run on an hardware accelerator, especially with the accelerator-compatible environments of \brax \citep{freeman2021brax}. They can be found in the library QDax~\citep{chalumeau2023qdax}.

\begin{table}[h!]
    \centering
    \caption{Identification of components shared across methods. 
    \label{tab:recap}}
    \begin{scriptsize}
    \begin{tabular}{lcccccc}
    \toprule
    & \textbf{Algorithm} & \textbf{Genetic Mutation} & \textbf{Map-Elites-based} & \textbf{PG update} & \textbf{SAC-based} & \textbf{TD3-based} \\
    \midrule
    \multirow{4}{*}{\rotatebox[origin=c]{90}{RL}}
    & \diaynsmerl & X & X & \checkmark & \checkmark & X\\
    & \dadssmerl & X & X & \checkmark & \checkmark & X\\
    & \dadssum & X & X & \checkmark & \checkmark & X\\
    & \diaynsum & X & X & \checkmark & \checkmark & X\\
    \midrule
    \multirow{4}{*}{\rotatebox[origin=c]{90}{QD}}
    & \pgame & \checkmark & \checkmark & \checkmark & X & \checkmark\\
    & \pgaaurora & \checkmark & \checkmark & \checkmark & X & \checkmark\\
    & \me & \checkmark & \checkmark & X & X & X\\
    & \aurora & \checkmark & \checkmark & X & X & X\\
    \bottomrule

    \vspace*{0.08cm}
    \end{tabular}
    
    \begin{tabular}{lccccc}
    \toprule
    & \textbf{Algorithm} & \textbf{Unsupervised diversity} & \textbf{SMERL sum.} & \textbf{DADS diversity} & \textbf{DIAYN diversity} \\
    \midrule
    \multirow{4}{*}{\rotatebox[origin=c]{90}{RL}}
    & \diaynsmerl & \checkmark  & \checkmark & X & \checkmark\\
    & \dadssmerl & \checkmark  & \checkmark & \checkmark & X\\
    & \dadssum & \checkmark  & X & \checkmark & X\\
    & \diaynsum & \checkmark  & X & X & \checkmark\\
    \midrule
    \multirow{4}{*}{\rotatebox[origin=c]{90}{QD}}
    & \pgame & X  & X & X & X\\
    & \pgaaurora & \checkmark  & X & X & X\\
    & \me & X & X & X & X\\
    & \aurora & \checkmark & X & X & X\\
    \bottomrule
    
    \end{tabular}
    
    \end{scriptsize}
\end{table}

\subsection{Algorithms under study}
\label{subsection:implem_algo}

\paragraph{\textbf{MAP-Elites}}

Our implementation of \me takes advantage of the just-in-time compilation and vectorization capacity of \jax, similarly to the implementation presented in \qdax \citep{lim2022accelerated}.
All environments are implemented in \brax (itself constructed on \jax), which enables to simulate multiple copies of the same environment in parallel, on the same device. Hence, we can simultaneously unroll thousands of episodes and evaluate as many policies. In addition, we put focus on implementing the mutation and insertion steps of \me to be compatible as well with the just-in-time compilation paradigm of \jax.  These design choices enable parallelization and lead to an overall speedup of several orders of magnitude compared to previous sequential implementations. Finally, we opt for the Iso+LineDD variation \citep{vassiliades2018discovering} for the updates of the population.

\paragraph{\textbf{PGA-MAP-Elites}} 

We provide here the first implementation of \pgame in \jax. This implementation benefits from the same vectorization strategies as \me which yields similar speedups. The major difference between these two algorithms lies in the policy variation step: rather than applying the Iso+LineDD mutation all the time, \pgame uses a mix of the latter variation and a policy gradient (\pgr) variation. As in \citet{nilsson2021policy}, this \pgr variation consists in updating the selected policies to follow a \tddd critic trained with transitions from a shared replay buffer.


\paragraph{\textbf{DIAYN}}

Our implementation closely follows the original implementation of \citet{eysenbach2018diversity}. For the latent skill variable, we rely on a categorical prior with dictionary of size $|\mathcal{Z}|$ and we sample $z$ with probability $p(z)=\frac{1}{|\mathcal{Z}|}$. For a given transition $(s,z)$ from the replay buffer, the reward is computed online as follows:
\begin{align}
    r_t = \log(q(z|s)) - \log(p(z))
\end{align}
The discriminator $q(z|s)$ is parameterized as a neural network that outputs a softmax distribution and is trained as a classifier to predict the skill $z$ from the current state $s$. The discriminator is updated at the same frequency as the policy and the critic. However, one distinction with the original implementation of \diayn stems from the way we take advantage of the \jax's parallelization capacity: we sample multiple skills and simultaneously generate trajectories by unrolling the skill-conditioned policy on identical copies of the environment. This way, we collect data from different skills at the same time, and accelerate the training of the discriminator. 

\paragraph{\textbf{DIAYN+REWARD and SMERL(DIAYN)}} \diaynsum is a supervised \rl method that consists in incorporating signal from the environment into the \diayn algorithm. At each timestep, we sum the environment reward and the reward of \diayn (multiplied by a scaling coefficient), see \equationautorefname~\ref{eq:diayn_sum}. Thus, we expect to discover diverse and high-performing skills. \diaynsmerl rather formulates the problem in a constrained manner: diversity should be maximized only when a policy is $\epsilon$-close to the return of an \textit{optimal} policy, $\epsilon$ being a margin set as an hyperparameter. To implement this constraint, we maintain for each transition in the replay buffer the return of its associated episode. Only transitions from an $\epsilon$-optimal episode are updated to add the diversity reward.

\paragraph{\textbf{DADS}}
Our implementation of \dads builds on the same basis as the original implementation. The prior chosen for $z$ is the same as in \diayn and the reward is defined as follows:
\begin{align}
    r_t = \log(q(s_{t+1}|s_t,z)) - \log(p(s))
\end{align}
where $p(s)$ is obtained by marginalizing over the (discrete) distribution of $z$. Following the work from \citet{sharma2019dynamics}, we define the skill-dynamics  model as a neural network which outputs a Gaussian Mixture distribution, with 4 experts and an identity covariance matrix. We also predict $\Delta s = s_{t+1} - s_t$ instead of $s_{t+1}$ and we exclude $s_t$ from the input, resulting in a skill-dynamics network of the form  $q(\Delta s| z)$. We maintain the running average and the standard deviation of the target $\Delta s$ to normalize the output of the skill-dynamics network. Our implementation trains the skill-dynamics network offline, similarly to the training of \diayn's discriminator, with parameters updated at the same frequency as the policy and the critic. Finally we take advantage of the batching capability of \jax along the same lines as for \diayn.

\paragraph{\textbf{DADS+REWARD and SMERL(DADS)}}
The summation method of \diaynsum (resp. \diaynsum) can be easily extended with any \mi method that provides a diversity reward. It naturally leads to the method \dadssum (resp. \dadssmerl).

\paragraph{\textbf{AURORA}}
\aurora is \qd algorithm introduced in \citet{cully2019aurora}. In this algorithm, the population is stored in an unstructured archive instead of a structured grid similarly to what is done in \me. The main difference of \aurora compared to usual \qd methods is that the behavior descriptor is learned instead of being manually defined by the user. To do so, an auto-encoder learns to encode and to reconstruct trajectories collected in the environment. The latent space defines the behavior space. In order to determine the behavior descriptor of a policy, its trajectory in the environment is given as input to the encoder. The output of the encoder is the behavior descriptor of the policy. The auto-encoder is trained following a simple geometric schedule (with decreasing frequency). At each training step, the auto-encoder learns to reconstruct the trajectories of the policies that are stored in the archive. Then, the behavior descriptors are computed with the new weights of the encoder and the unstructured archive is updated accordingly. The genetic mechanism used to generate offspring in our study is the same as \me: we use the Iso+LineDD variation.

\paragraph{\textbf{PGA-AURORA}}
\pgaaurora is a new algorithm that we introduce in this paper. It has components inspired by \aurora and \pgame. It uses the unsupervised behavior descriptor definition from \aurora, as well as its unstructured archive. But instead of using purely genetic mutations like \aurora, \pgaaurora uses the policy-gradient mutation introduced in \pgame \citep{nilsson2021policy}. This baseline is interesting because it has exactly the same prior knowledge as the \mirl methods considered in this benchmark: it does not rely on the definition of a behavior descriptor but a prior can be injected. In our experiments, the population of the unstructured archive is the same as for the other \qd methods and the latent space size is set to 5. The auto-encoder uses LSTM units to encode and decode the given trajectories. The implementation of \pgaaurora is also fully implemented in \jax.

\subsection{Passive repertoire: measuring diversity for latent-conditioned policies}
Skill-discovery \rl methods considered in this paper do not maintain a repertoire to store diverse solutions. Instead, policies are indexed by the possible values that the discrete latent variable can take for \diayn, \dads, and \smerl. For these methods, a possible approach to introduce the same type of repertoire used by \qd methods is to fill one in an ad-hoc fashion (i.e. post training) by enumerating all possible values for the latent variable and evaluating the policy conditioned on this value each time. However, this puts these methods at a clear disadvantage if the cardinality of the latent space is small compared to the size of the repertoire given that most \qd metrics are increasing functions of the number of non-empty cells. We could increase the size of the discrete latent space to match the size of the repertoire but skill-discovery \rl methods become unstable when the size of the latent space is large \citep{eysenbach2018diversity}. Instead, we choose to fill the repertoire during training with all the intermediate policies found by skill-discovery \rl methods, in the same fashion as what is done internally in \qd methods. Specifically, every 100,000 training step, we evaluate the fitness and behavior descriptors of all of the latent-conditioned policies and insert them in the repertoire whenever possible, following the same insertion rules as \me : if the cell corresponding to the behavior descriptor value is empty or if the solution stored previously in this cell has lower fitness. We stress that, in our experiments, the \qd methods make use of the same behavior descriptors used to compute the \qd metrics, which induces a bias in the analysis. To make the study less biased, we embed the choice of the behavior descriptor space into skill-discovery \rl methods whenever possible. This is possible for \omnidir environments where we can choose a discriminator function for all skill-discovery methods that is conditioned on the behavior descriptor as the latter is just a function of the current state. This is however not possible for \unidir environments.

\subsection{Details on QD-score computation}
As commonly done in the \qd literature, we add an offset to the fitnesses when computing the \qd score to guarantee that it is an increasing function of the coverage. The repertoires, which are initialized identically across methods, are of size 1024 and are constructed using Centroidal Voronoi Tessellations \citep{vassiliades2017using}.

\subsection{Environments and modifications}

To define our environments, we use the \brax physics simulator, which is provided in an open-source package fully implemented in \jax \cite{freeman2021brax}. \brax builds environments from configurations that describe the elements to simulate. Specifically, the properties of the body parts of the agent, its joints, and its actuators, as well as the types of contacts that need to be taken into account during the simulations and a set of physic constants (including friction and gravity) need to be specified in the configuration.

Implementations for the base environments mentioned in section~\ref{sec:domains}, (i.e. \ant, \walker, and \hc) are provided in the \brax package. The tasks derived from these environments (i.e. \unidir, \antomni, \antmaze, \anttrap) can be implemented with minor code changes. In \unidir, we pass along the feet/ground contact information from the simulator to the user to be able to compute the behavior descriptors. In \antomni, we do the same for the (x, y) position of the center of gravity of the robot. In \anttrap, three walls are added in front of the ant to create the trap and the (x, y) position of the center of gravity is made available. Similarly, we have created a maze with simulated walls to generate the \antmaze task.

In \ptmaze, actions are bounded 2-dimensional vectors corresponding to the $(x, y)$ increments that can be added to the current position. The observation received at each timestep is the current point's position on the 2D plane. The reward signal is defined as the negative euclidean distance between the current position and the target position.

In \anttrap, the reward signal is defined as the x-axis velocity minus the energy consumption. In \antmaze, the reward signal is defined as the negative euclidean distance between the ant's center of gravity and the target position.

For adaption experiments, we took inspiration from both the \qd and the \rl literature. The experiment consisting in introducing leg damage has been studied in \citet{cully2015robots} and \citet{kumar2020one}. Experiments that involve changing the gravity property have been used in \citet{gaya2021learning}. A major difference in this work lies in the fact that we study a significantly larger range of perturbations in each setting, which enables us to get a sense of the adaptation capabilities of each method against both small and significant perturbations. The adaptation task for \antmaze is also inspired from prior work in the \qd literature \citep{pierrot2022diversity}.

In order to create variations of the environment in our adaptation tasks, we apply wrappers to modify a specific entry of the configuration. For instance, to change the gravity value, we (1) retrieve the configuration of the environment, (2) fetch the value of the gravity constant, (3) multiply it by a given input and (4) set the new value in the configuration. Similarly, we change the behavior of the actuator of a specific joint by updating the "actuator\_strength" entry in the \brax configuration. In order to create the leg-dysfunction adaptation task, we pick a multiplicative factor and multiply the "actuator\_strength" of all the joints that link bodies of the concerned leg by this factor. A given action input will result in a different update of the joint angle in the resulting physical simulation compared to the original simulation. At the lower extreme of the range (i.e. 0), this makes the actuator completely ineffective. At the higher extreme of the range (i.e. 4.5), the joint swings much faster than in the nominal case for a given input command.

Below we detail several elements to change in the \brax configurations in order to replicate our adaptation tasks. We also provide the code to replicate those.
\begin{itemize}
    \item Leg dysfunction in \walkeruni: the leg impacted by the dysfunction is the left leg of the \walker. The impacted joints in \brax configuration are called "thigh\_left\_joint", "leg\_left\_joint" and "foot\_left\_joint".
    \item Leg dysfunction in \antuni: the leg impacted by the dysfunction in the ant corresponds to the following joints in \brax configuration: "\$ Torso\_Aux 4" and "Aux 4\_\$ Body 13".
    \item Leg dysfunction in \hcuni: the leg impacted by the dysfunction is the front leg of the \hc. The impacted joints in the \brax configuration are called "fthigh", "fshin" and "ffoot".
    \item Gravity update in all the \unidir tasks: the gravity parameter of the environment is usually -9.8 by default. For each environment, we multiply it by a range of 20 values, to get values ranging from -2.45 (low gravity) to -490 (high gravity).
    \item Target position update in \antmaze: we sampled 10 new positions uniformly inside the maze (i.e. in $[-3, 38]^2$). The values used in our final experiments are reported on Table~\ref{tab:new_target_pos}.
\end{itemize}

For the hierarchical learning task, we have added the $(x, y)$-position of the robot to the observation because it is needed by the meta-controller to take relevant actions. 

\begin{table}[ht!]
    \begin{center}
    \begin{tabular}{ccc} 
        \toprule
        \textbf{Run number} & \textbf{x position} & \textbf{y position} \\
        \midrule
        Training & $35.0$ & $0.0$ \\
        \midrule
        Adaptation 1 & $11.74$ & $35.16$ \\
        Adaptation 2 & $-2.33$ & $14.86$ \\
        Adaptation 3 & $32.48$ & $32.37$ \\
        Adaptation 4 & $29.52$ & $17.46$ \\
        Adaptation 5 & $12.46$ & $7.05$ \\
        Adaptation 6 & $17.36$ & $4.45$ \\
        Adaptation 7 & $8.38$ & $25.17$ \\
        Adaptation 8 & $24.02$ & $34.36$ \\
        Adaptation 9 & $21.75$ & $14.12$ \\
        Adaptation 10 & $19.51$ & $14.58$ \\
    
        \bottomrule
    
    \end{tabular}
    \end{center}
    \caption{New target positions in \antmaze adaptation task. During the training phase, the position is set to (35, 0) but varies during adaptation. The 10 different values used during our adaptation experiments are reported.}
    \label{tab:new_target_pos}
\end{table}

For the \hchurdles, we add obstacles in front of the halfcheetah in the x-direction. To do so, we add capsule body to the Brax simulation configuration and include collisions between the robot's body and the capsules. Those hurdles are placed every three meters, have a length (y-axis) of 1 meter, a height (z-axis) of $0.25$ meter and a radius of $0.25$ meter.

\section{Hyperparameters}
\label{sec:hyperparams}
In this section, we document the various hyperparameters that have been used in the study and the selection process that led us to pick these values.

\begin{table}[ht!]
    \begin{center}
    \begin{tabular}{lc}
        \toprule
        \textbf{Hyperparameter} & \textbf{Value} \\
        \midrule
        Policy learning rate &  $0.0003$\\
        Critic learning rate &  $0.0003$\\
        Environment batch size & $200$ \\
        Batch size & $256$ \\
        Discount factor &  $0.99$\\
        Entropy coefficient & $0.1$ \\
        Policy hidden layers size &  $[256,256]$\\
        Critic hidden layers size &  $[256,256]$\\
       \midrule
          \multicolumn{2}{c}{\diayn} \\
        \midrule
        Discriminator learning rate &  $0.0003$\\
        Discriminator hidden layers size &  $[256,256]$\\
        Number of discrete skills &  $5$\\
        
       \midrule
          \multicolumn{2}{c}{\dads} \\
        \midrule
        
        Skill-dynamics learning rate &  $0.0003$\\
        Skill-dynamics hidden layers size &  $[256,256]$\\
        Number of discrete skills &  $5$\\
        \bottomrule
        
    \end{tabular}
    \end{center}
    \caption{Hyperparameters for \diayn and \dads. Most parameters are inspired from the original papers that introduced the methods.}
    \label{table:hp_sac}
\end{table}

\paragraph{Skill-discovery MI RL methods.}

Throughout our evaluation, we use the same hyperparameter values for shared parameters across all methods and environments. Table~\ref{table:hp_sac} details the choice of hyperparameters for \diayn and \dads, which both use \sac internally. This set of hyperparameters values, derived from \citet{osa2021discovering} and \citet{kumar2020one}, yields good performance across all environments for both methods. However, some environment-dependent parameters require more care. In particular, the global reward multiplier of the environment is set to $1.0$ for all environments except for \hcuni (resp. \antuni), where it is set to $5.0$ (resp. $10.0$), following observations reported in \brax \citep{freeman2021brax}. One crucial parameter choice for \diaynsum and \dadssum (and by extension \diaynsmerl and \dadssmerl) is the \textbf{diversity reward scale}, $\beta$, that affects the diversity term in the weighted sum:
\begin{align}
    r_t = r_t^\text{environment} + \beta r_t^\text{diversity}
    \label{eq:diayn_sum}
\end{align}
We set $\beta=2.0$ as it has proved to favor the emergence of diverse and high-performing skills over all environments in our experiments, setting aside \antmaze, where we set $\beta=3.0$ to adjust to the larger scale of the environment's rewards, and \antomni, where we set $\beta=4.0$ to compensate for the fact that the environment's reward drives the agent to learn static strategies by penalizing the use of the motors. The same values of $\beta$ were used for \smerl. To select the \textit{optimal} \smerl return, similarly to \citet{kumar2020one}, we evaluate a \sac agent on 5 seeds and select the median of the best-return policy as the target. The margin-to-optimal return is set as $10\%$ of the target return. For all \diayn and \dads variants, we select a discrete skill distribution of size $|\mathcal{Z}|=5$, following choice made in \smerl \citep{kumar2020one}. It is often hard to learn bigger numbers of distinct skills \citep{eysenbach2018diversity}.

Finally, the value for environment batch size was set to 200. To select it, we tried a range of values similarly to what was done in \brax paper \citep{freeman2021brax}. In the end, 200 was the highest value that sped up the training as much as possible while not perturbing the learning dynamics.

\begin{wraptable}{r}{0.5\textwidth}
    \begin{center}
    \begin{tabular}{lc}
        \toprule
        \textbf{Hyperparameter} & \textbf{Value} \\
        \midrule
        Environment batch size & $1000$ \\
        Policy hidden layers size & $[256,256]$\\
        Iso sigma &  $0.005$ \\
        Line sigma &  $0.05$ \\
        \bottomrule
    
    \end{tabular}
    \end{center}
    \caption{Hyperparameters for \me.}
    \label{table:hp_mapelites}
\end{wraptable}

\begin{table}[h!]
    \begin{center}
    \begin{tabular}{lc} 
        \toprule
        \textbf{Hyper-parameter} & \textbf{Value} \\
        \midrule
        Environment batch size & $100$ \\
        Policy learning rate &  $0.001$\\
        Critic learning rate &  $0.0003$\\
        Policy hidden layers size &  $[256,256]$\\
        Critic hidden layers size &  $[256,256]$\\
        
        Policy noise & $0.2$\\
        Noise clip & $0.5$\\
        Discount & $0.99$\\
        Reward scaling & $1.0$\\
        
        Policy gradient proportion & $50\%$ \\
        Critic training steps & $300$\\
        Policy training steps & $100$\\
        Iso sigma & $0.005$ \\
        Line sigma & $0.05$ \\
        \bottomrule
    
    \end{tabular}
    \end{center}
    
    \caption{Hyperparameters for \pgame. Most parameters values are standard in the literature on \me and \tddd for the tasks considered in this paper. For fairness, we used the same architecture for all methods under study whenever possible. The number of policy gradient steps is increased compared to the original implementation as this has proved to improve performance.}
    \label{table:hp_pgame}
\end{table}

\paragraph{QD methods}

For \pgame, we took the hyperparameter values used in the original implementation \citep{nilsson2021policy}, see Table~\ref{table:hp_pgame}, setting aside the number of \pgr steps applied to a policy when it is updated by a policy gradient variation. Instead of the value $10$ used in the original paper, we opted for a value of $100$ which yielded better performances in practice in our experiments.

Note that we pick a different value for the environment batch size for \me and \pgame. This hyperparameter corresponds to the number of policies that are evolved and evaluated in parallel at each step of the algorithm. For \me, empirical evidence from a previous study, namely \cite{lim2022accelerated}, suggests that a large environment batch size dramatically speeds up the training phase without impacting the performance. Hence we chose to batch $1000$ environments in parallel in order to leverage as much as possible the vectorization capabilities of \jax. 

\begin{wraptable}{r}{0.5\textwidth}
    \begin{center}
    \begin{tabular}{lc}
        \toprule
        \textbf{Hyperparameter} & \textbf{Value} \\
        \midrule
        Latent space size & $5$ \\
        Archive initial l value & $0.2$\\
        \bottomrule
    
    \end{tabular}
    \end{center}
    \caption{\aurora-specific hyperparameters for \aurora. The other hyperparameters are similar to \me.}
    \label{table:hp_mapelites}
\end{wraptable}

However, there is no such empirical evidence for \pgame. Specifically, since some elements are shared across the policies trained in parallel (e.g. the replay buffer of the underlying \tddd), scaling up the environment batch size may negatively impact the training. As a matter of fact, multiplying by $10$ the environment batch size for \pgame will not result in an $10$-fold increase of the number of environment steps collected in the same amount time (which would be the case for \me). For this reason, we decided to keep the exact same value of environment batch size as in the original \pgame study (i.e. $100$).

\clearpage

\section{Evolution of QD metrics during training}

\begin{figure}[h!]
    \centering
    \includegraphics[width=1.\textwidth]{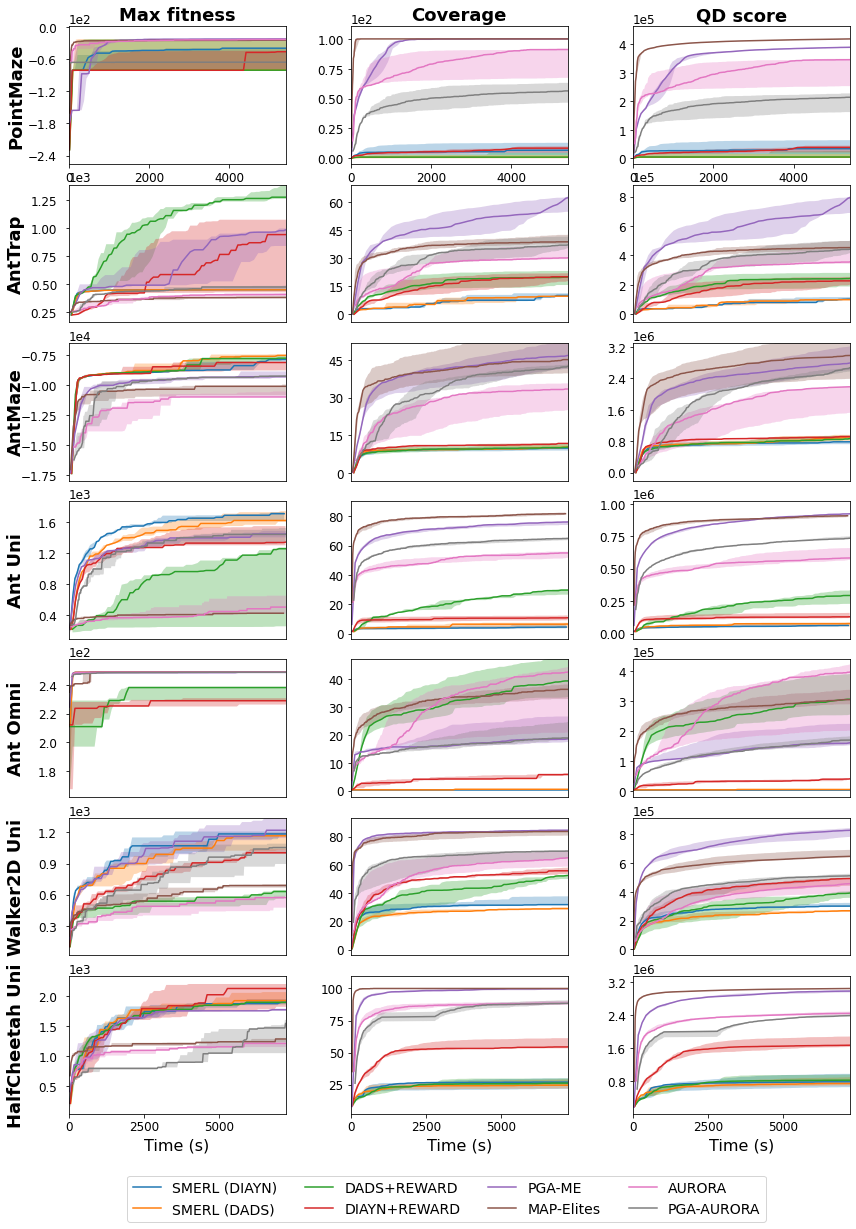}

    \caption{Evolution of the maximum fitness, coverage and \qd score during the training phase (2 hours), for all methods under study.}
    \label{fig:qd_metrics_time}
\end{figure}

In this section, we report some additional results which shed light on the evolution of performance during training w.r.t. the \qd metrics  described in Section \ref{sec: exp}. In the main body of this paper, the final performances (\qd score and maximum fitness) achieved by all methods after two hours of training are reported in Table~\ref{qd-score-comparison}  while visualizations of the best final grids obtained for each method are also available in Figure~\ref{fig:repertoires}. In this section, we plot the evolution of the \qd metrics during training for all methods in \figureautorefname~\ref{fig:qd_metrics_time}. Additionally, Table~\ref{table:training_step} reports the number of environment steps carried out during training, averaged across the 8 evaluated environments, for all methods.

\begin{wraptable}{r}{0.55\textwidth}
    \begin{center}
    \begin{tabular}{lc}
        
        \toprule
        \textbf{Algorithm} & \textbf{Env. steps} \\
        \midrule
    
        \diaynsmerl & $1.0 \times 10^7$ \\
        \diaynsum & $1.0 \times 10^7$ \\
        \dadssmerl & $8.5 \times10^6$ \\
        \dadssum & $8.5\times 10^6$ \\
        \pgame & $1.8\times 10^8$ \\
        \me & $1.6 \times10^9$ \\
        \aurora & $1.2 \times10^9$ \\
        \pgaaurora & $1.1 \times 10^8$\\
        \bottomrule
    
    \end{tabular}
    \end{center}
    \caption{Number of training steps carried out (on average) during training by the various methods under study.}
    \label{table:training_step}
\end{wraptable}

In our study, we chose to express the training budget in terms of time instead of environment steps. As hardware accelerators are getting more efficient, we believe that the intrinsic capacity of an algorithm to make the most of the available hardware is an important property to assess. Furthermore, with the development of highly accurate simulators, many real-world applications can be simulated, which enables to break down the training phase in two steps: (1) training in simulation (where the number of environment interactions is irrelevant, only the time budget is), and (2) sim-to-real training; rather than a single training phase in the real world.

\begin{figure}[h!]
    \centering
    \includegraphics[width=1.\textwidth]{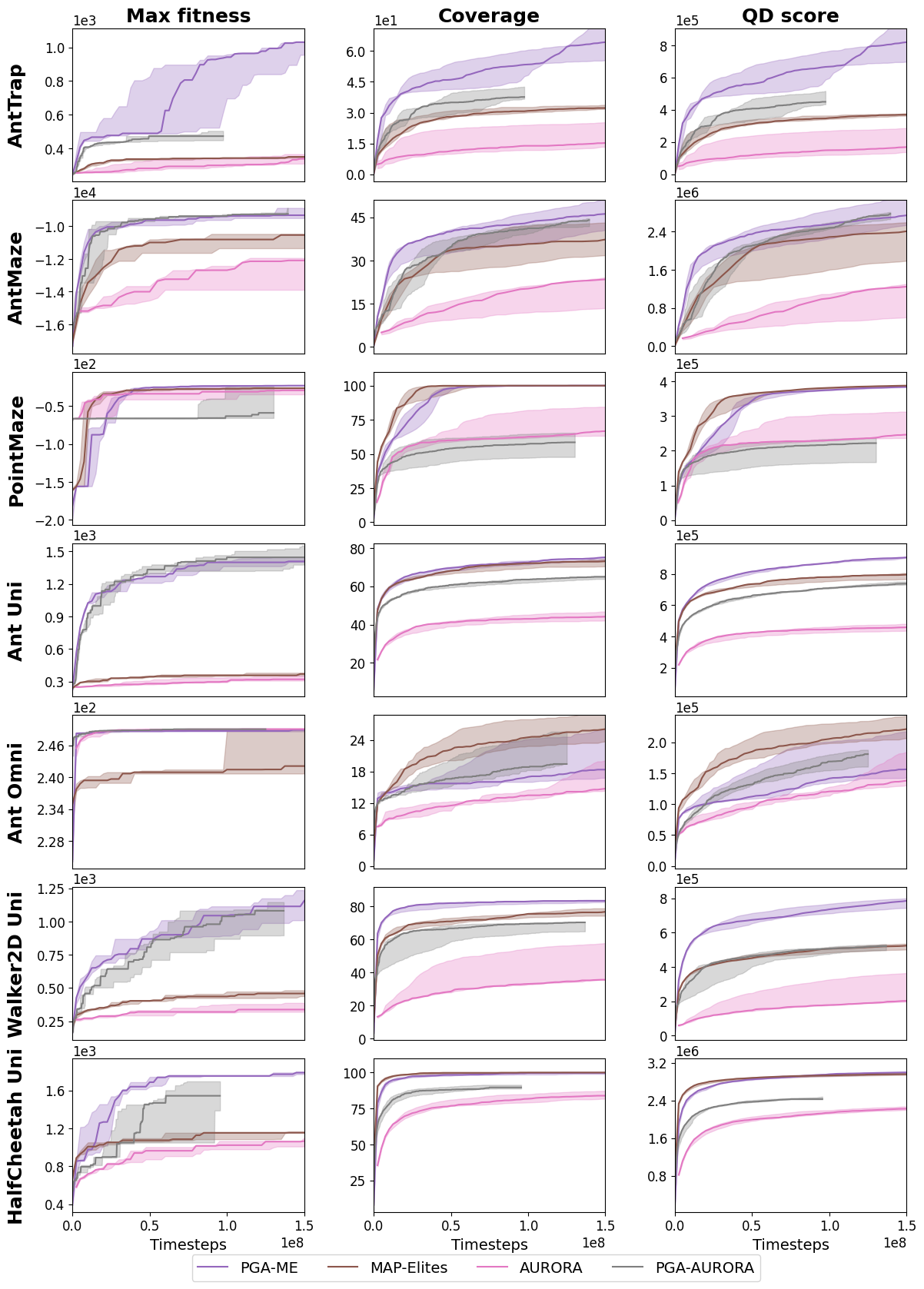}

    \caption{Evolution of the maximum fitness, coverage and \qd score along environment interactions, during the training phase. Comparing \qd algorithms \me, \aurora, \pgame and \pgaaurora on $1.5\times10^8$ timesteps.}
    \label{fig:qd_metrics_evo_step}
    
\end{figure}

\begin{figure}[h!]
    \centering
    \includegraphics[width=1.\textwidth]{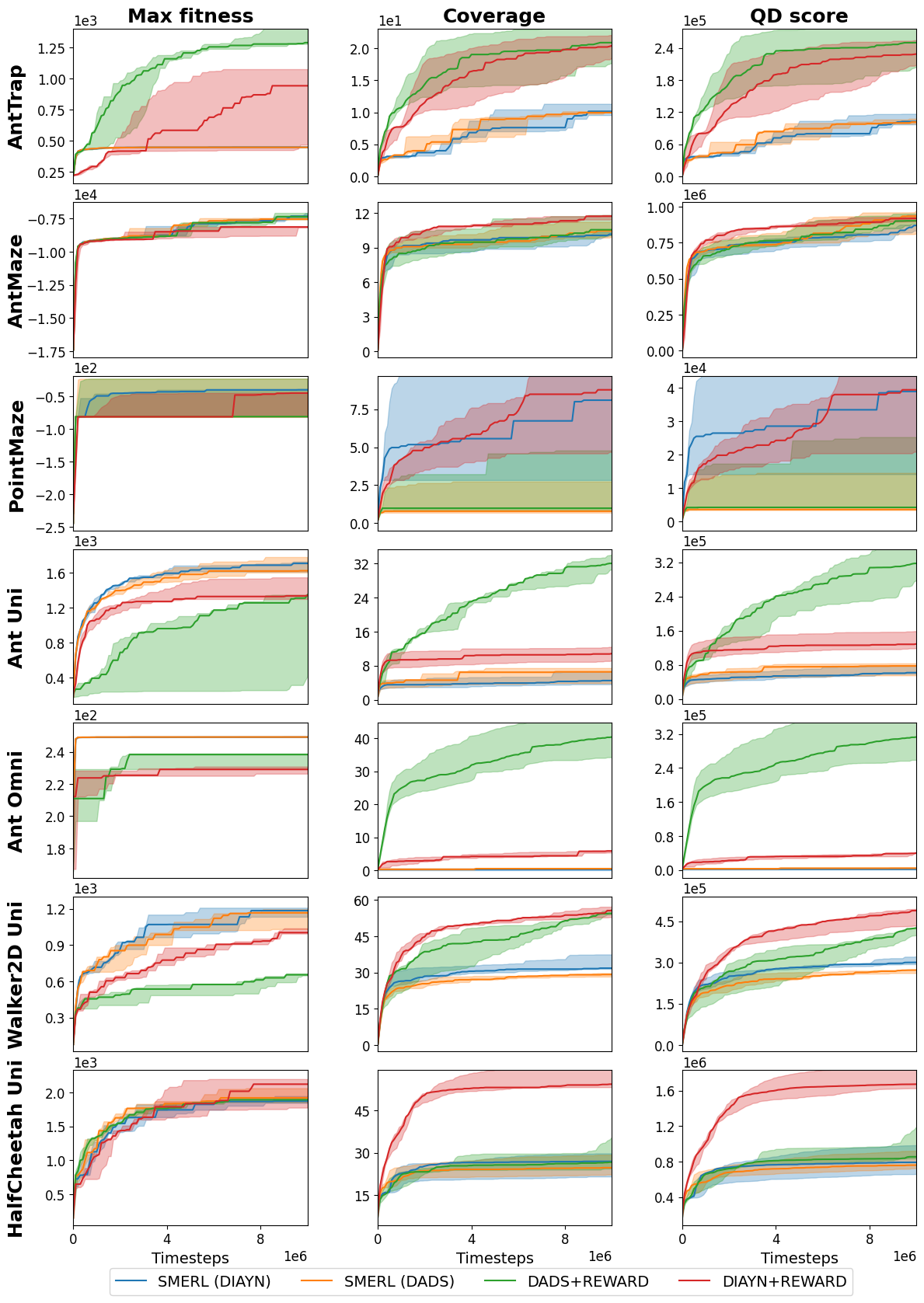}
    \caption{Evolution of the maximum fitness, coverage and \qd score along environment interactions, during the training phase. Comparing skill-discovery algorithms \diaynsum, \diaynsmerl, \dadssum and \dadssmerl  on $10^7$ timesteps.}
    \label{fig:qd_metrics_rl_step}
\end{figure}

Since sample-efficiency is critical for (2), we also provide plots showing the evolution of the \qd metrics w.r.t. the number of environment interactions during training. We plot the evolution of \qd metrics for all \qd (resp. skill-discovery) methods as a function of the number of environment steps, with a maximum horizon of $1.5 \times 10^8$ (resp. $10^7$) steps, on \figureautorefname~\ref{fig:qd_metrics_evo_step} (resp. \figureautorefname~\ref{fig:qd_metrics_rl_step}). \figureautorefname~\ref{fig:qd_metrics_evo_step} highlights the gain in sample efficiency yielded by the \pgr variation in \pgame compared to \me (although \me performs well on a time scale, see \figureautorefname~\ref{fig:qd_metrics_time}).

We also provide the standard deviation of the maximum fitness and \qd-score reported in Table~\ref{tab:max_fitness} in Table~\ref{tab:max_fitness_std}.

\begin{table}[b]
  \caption{Maximum fitness and \qd score at the end of the training phase (median over 5 seeds). The standard deviation is reported.}
  \centering
  \resizebox{.98\textwidth}{!}{
  \centering

  \begin{tabular}{lccccccc}
    \toprule
         & \anttrap & \antmaze & \ptmaze & \antuni & \antomni & \walkeruni & \hcuni \\
    \midrule
        \multicolumn{8}{c}{\textbf{Maximum fitness}} \\
    \midrule
    \diaynsmerl & 4.49 ($\pm$ 0.01)\text{e}2 & -7.88 ($\pm$ 0.41)\text{e}3 & -4.00 ($\pm$2.57)\text{e}1 & \textbf{1.71 ($\pm$0.09)e3} & \textbf{2.49 ($\pm$0.00)e2} & 1.19 ($\pm$0.06)\text{e}3 & 1.89 ($\pm$0.14)\text{e}3 \\
    \dadssmerl & 4.48 ($\pm$0.02)\text{e}2 & \textbf{-7.62 ($\pm$0.43)e3} & -8.10 ($\pm$3.19)\text{e}1 & 1.61 ($\pm$0.18)\text{e}3 & \textbf{2.49 ($\pm$0.00)e2} & 1.17($\pm$0.14)\text{e}3 & 1.92 ($\pm$0.21)\text{e}3 \\
    \dadssum & \textbf{1.28 ($\pm$0.41)e3} & -7.81 ($\pm$0.49)\text{e}3 & -8.10 ($\pm$3.20)\text{e}1 & 1.24 ($\pm$0.52)\text{e}3 & 2.38 ($\pm$0.11)\text{e}2 & 6.30 ($\pm$0.48)e2 & 1.89 ($\pm$0.03)\text{e}3 \\
    \diaynsum & 9.42 ($\pm$3.43)\text{e}2 & -8.13 ($\pm$0.51)\text{e}3 & -4.50 ($\pm$2.53)\text{e}1 & 1.34 ($\pm$0.21)\text{e}3 & 2.29 ($\pm$0.04)\text{e}2 & 1.00 ($\pm$0.13)\text{e}3 & \textbf{2.13 ($\pm$0.27)e3} \\
    \pgame & 9.77 ($\pm$2.22)\text{e}2 & -9.32 ($\pm$0.43)\text{e}3 & \textbf{-2.30 ($\pm$0.03)e1}  & 1.45 ($\pm$0.06)\text{e}3 & \textbf{2.49 ($\pm$0.00)e2} & \textbf{1.22 ($\pm$0.14)e3} & 1.75 ($\pm$0.04)e3 \\
    \pgaaurora & 4.72 ($\pm$0.31)\text{e}2 & -9.27 ($\pm$0.06)\text{e}3 & -6.3 ($\pm$2.3)\text{e}1 & 1.44 ($\pm$0.19)\text{e}3  & \textbf{2.49 ($\pm$0.00)e2} & 1.05 ($\pm$0.16)\text{e}3 & 1.47 ($\pm$0.29)\text{e}3 \\
    \me & 3.81 ($\pm$0.11)\text{e}2 & -10.1 ($\pm$0.05)\text{e}4 & -2.5 ($\pm$0.05)\text{e}1 & 4.25 ($\pm$0.26)\text{e}2  & \textbf{2.49 ($\pm$0.00)e2}& 6.87 ($\pm$0.70)\text{e}2 & 1.23 ($\pm$0.06)\text{e}3 \\
    \aurora & 4.05 ($\pm$0.52)\text{e}2 & -1.10 ($\pm$0.09)\text{e}4 & -2.35 ($\pm$0.12)\text{e}1 & 5.05 ($\pm$0.26)\text{e}2  & \textbf{2.49 ($\pm$0.00)e2} & 5.71 ($\pm$1.34)\text{e}2 & 1.21 ($\pm$0.08)\text{e}3 \\
    \midrule
        \multicolumn{8}{c}{\textbf{QD score}} \\
    \midrule
    \diaynsmerl & 1.02 ($\pm$ 0.17)\text{e}5 & 7.84 ($\pm$ 0.44)\text{e}5 & 38.9 ($\pm$13.8)\text{e}4 & 6.20 ($\pm$ 1.38)\text{e}4 & 1.95 ($\pm$1.70)\text{e}3 & 3.01 ($\pm$0.20)\text{e}5 & 7.92 ($\pm$2.23)\text{e}5 \\
    \dadssmerl & 9.88 ($\pm$ 0.56)\text{e}4 & 8.22 ($\pm$ 0.07)\text{e}5 & 3.57 ($\pm$ 6.84)\text{e}3 & 0.83 ($\pm$ 1.17)\text{e}5  & 4.15 ($\pm$1.11)\text{e}3  & 2.69 ($\pm$0.38)\text{e}5& 7.51 ($\pm$1.60)\text{e}5 \\
    \dadssum & 2.42 ($\pm$ 0.80)\text{e}5 & 9.12 ($\pm$ 0.76)\text{e}5 & 0.42 ($\pm$ 7.56)\text{e}4 & 2.88 ($\pm$ 1.07)\text{e}5 & 3.05 ($\pm$1.22)\text{e}5 & 3.89 ($\pm$0.41)\text{e}5 & 8.31 ($\pm$3.91)\text{e}5\\
    \diaynsum & 2.27 ($\pm$ 0.62)\text{e}5 & 9.18 ($\pm$ 0.16)\text{e}5 & 4.04 ($\pm$ 1.60)\text{e}4 & 1.29 ($\pm$ 0.39)\text{e}5 & 4.00 ($\pm$0.39)\text{e}4 & 4.92 ($\pm$0.35)\text{e}5 & 1.68 ($\pm$0.18)\text{e}6\\
    \pgame & \textbf{7.89 ($\pm$2.87)e5} & 2.74 ($\pm$ 0.50)\text{e}6 & 3.92 ($\pm$ 0.03)\text{e}5 & \textbf{9.19($\pm$ 0.09)\text{e}5} & 1.58 ($\pm$0.47)\text{e}5 & \textbf{8.23 ($\pm$0.42)e5} & 2.98 ($\pm$0.52)\text{e}6 \\
    \pgaaurora & 4.40 ($\pm$0.64)\text{e}5 & 2.59 ($\pm$0.15)\text{e}6 & 2.88 ($\pm$0.80)\text{e}5 & 7.33 ($\pm$0.19)\text{e}5 & 1.69 ($\pm$0.42)\text{e}5 & 5.10 ($\pm$0.64)\text{e}5 & 2.39 ($\pm$0.06)\text{e}6 \\
    \me & 4.53 ($\pm$ 0.54)\text{e}5 & \textbf{2.98 ($\pm$0.68)\text{e}6} & 4.22 ($\pm$ 0.01)\text{e}5 & 9.08 ($\pm$0.20)\text{e}5 & 3.04 ($\pm$0.74)\text{e}5 & 6.45 ($\pm$0.33)\text{e}5 & \textbf{3.05 ($\pm$0.00)\text{e}6} \\
    \aurora & 3.54 ($\pm$0.11)\text{e}5 & 2.18 ($\pm$0.59)\text{e}6 & \textbf{4.58 ($\pm$0.72)\text{e}5} & 5.82 ($\pm$0.79)\text{e}5 & \textbf{4.00 ($\pm$1.33)\text{e}5} & 4.59 ($\pm$1.16)\text{e}5 & \textbf{2.45 ($\pm$0.04)\text{e}6} \\
    \bottomrule
  \end{tabular}
}
  \label{tab:max_fitness_std}
\end{table}

\newpage
\newpage
\newpage

\section{Additional visualizations of the training phase}

\begin{figure}[h!]
    \centering
        \includegraphics[height=0.7\textheight]{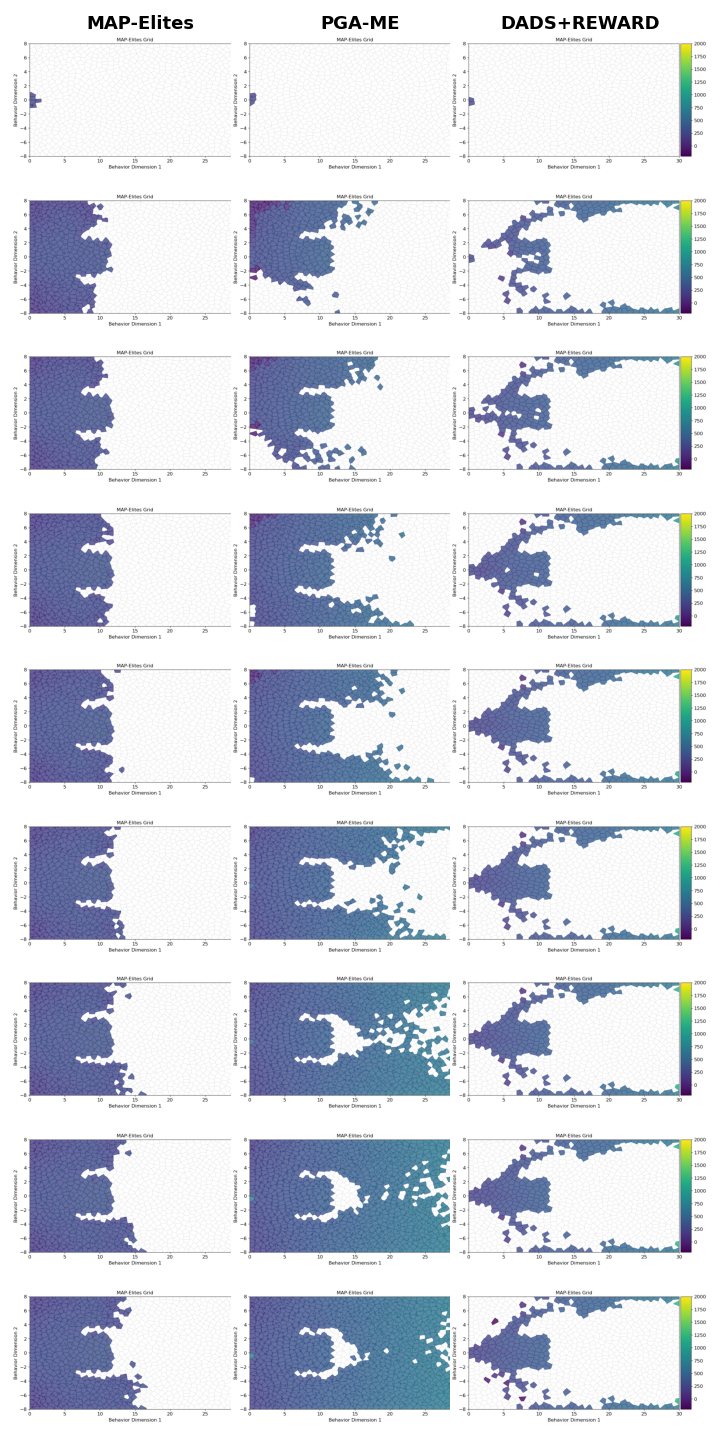}
    \caption{\textbf{Intermediate repertoires of MAP-Elites, PGA-MAP-Elites and DADS+REWARD on AntTrap.} Visualizations were generated at regular intervals from the start of training (top) to halfway through the training phase (i.e. at the 1h mark) (bottom).}
    \label{fig:grids_anttrap}
\end{figure}

In this section, we provide additional visualizations of the training phase. \figureautorefname~\ref{fig:grids_anttrap} display nine intermediate repertoires obtained by \me, \pgame and \dadssum at various points during training, evenly distributed over the first hour of training. These visualizations shed some light on the mechanisms through which these repertoires are filled during training. 

\begin{wrapfigure}{r}{0.45\textwidth}
    \centering
    \includegraphics[width=0.66\linewidth]{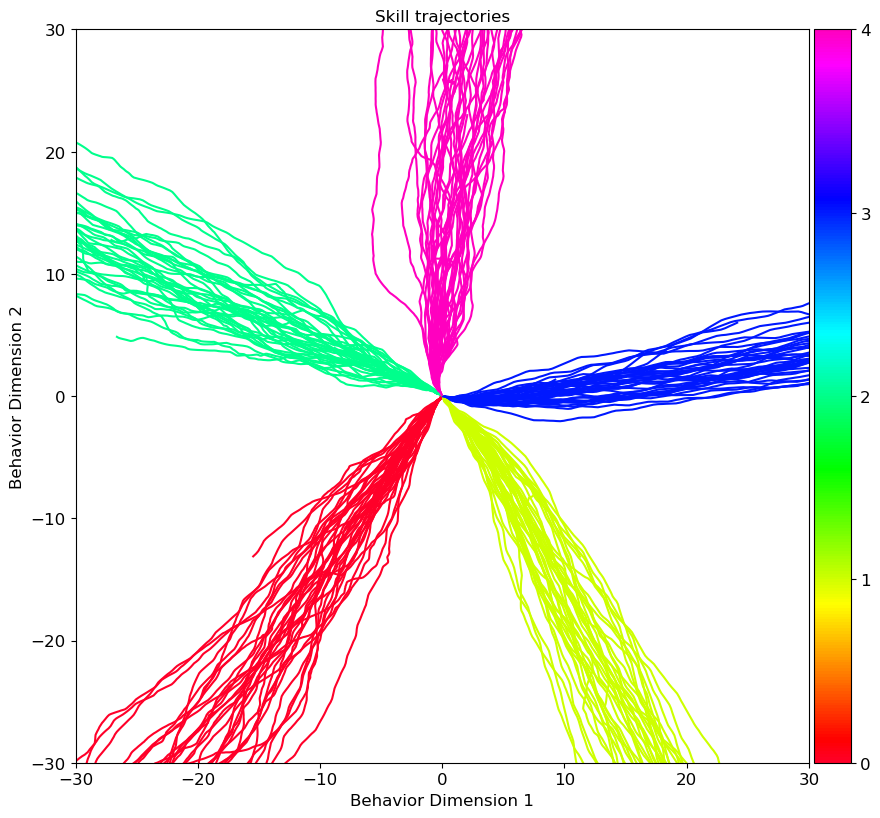}
    \caption{Visualization of final skills' trajectories (40 random seeds per skill) learned by \dadssum on \antomni.}
    \label{fig:dads_sum_skill_ant_omni}
\end{wrapfigure}

For instance, we observe that the use of a policy gradient in \pgame focuses the search toward solutions that move alongside the x-axis direction (recall that only x-axis movements are rewarded in the locomotion environments) whereas the first checkpoint for \me shows solutions that can go everywhere alongside the y-axis even though no solution has been able to hit the trap yet. \me struggles to further fill the repertoire (after 20M steps, the repertoire has almost reached its final state), whereas \pgame continuously improves the repertoire coverage to finally reach (nearly) 100\% after two hours of training, see \figureautorefname~\ref{fig:repertoires}. Finally, note that \dadssum rapidly reach high-returns behaviors and focus on improving them through the training. However, it neglects intermediate policies that are sub-optimal but diversity-carrying.

\begin{wrapfigure}{r}{0.45\textwidth}
    \begin{subfigure}{1\linewidth}
        \centering
        \includegraphics[width=0.7\linewidth]{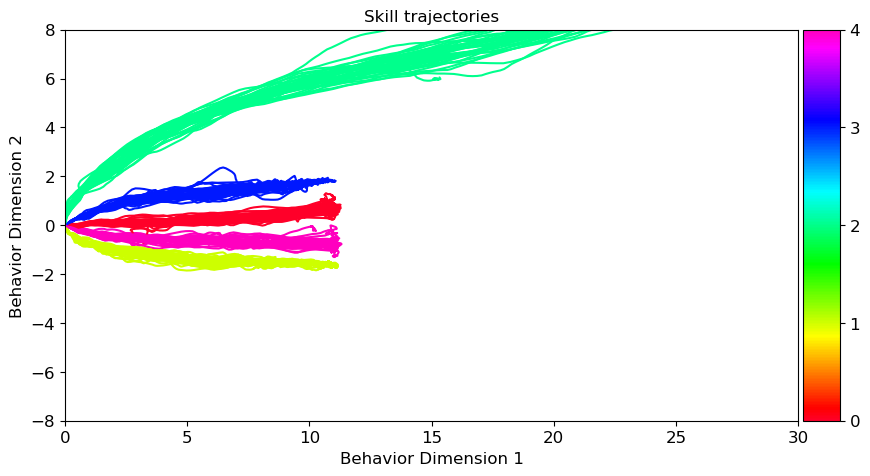}
    \end{subfigure}%
    \hfill
    \begin{subfigure}{1\linewidth}
        \centering
        \includegraphics[width=0.7\linewidth]{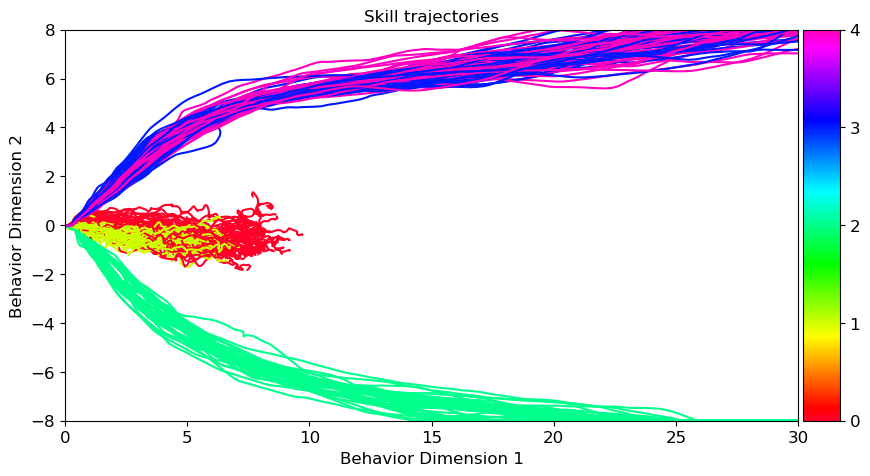}
    \end{subfigure}
    \caption{Visualizations of the final skills' trajectories (40 random seeds per skill) learned by \diaynsum (left) and \dadssum (right) on \anttrap.}
    \label{fig:diayn_dads_skills}
\end{wrapfigure}

\begin{figure}[h!]
     \begin{subfigure}{0.49\textwidth}
        \captionsetup{width=.85\linewidth}
         \hspace{-37pt}
         \includegraphics[height=0.8\textheight]{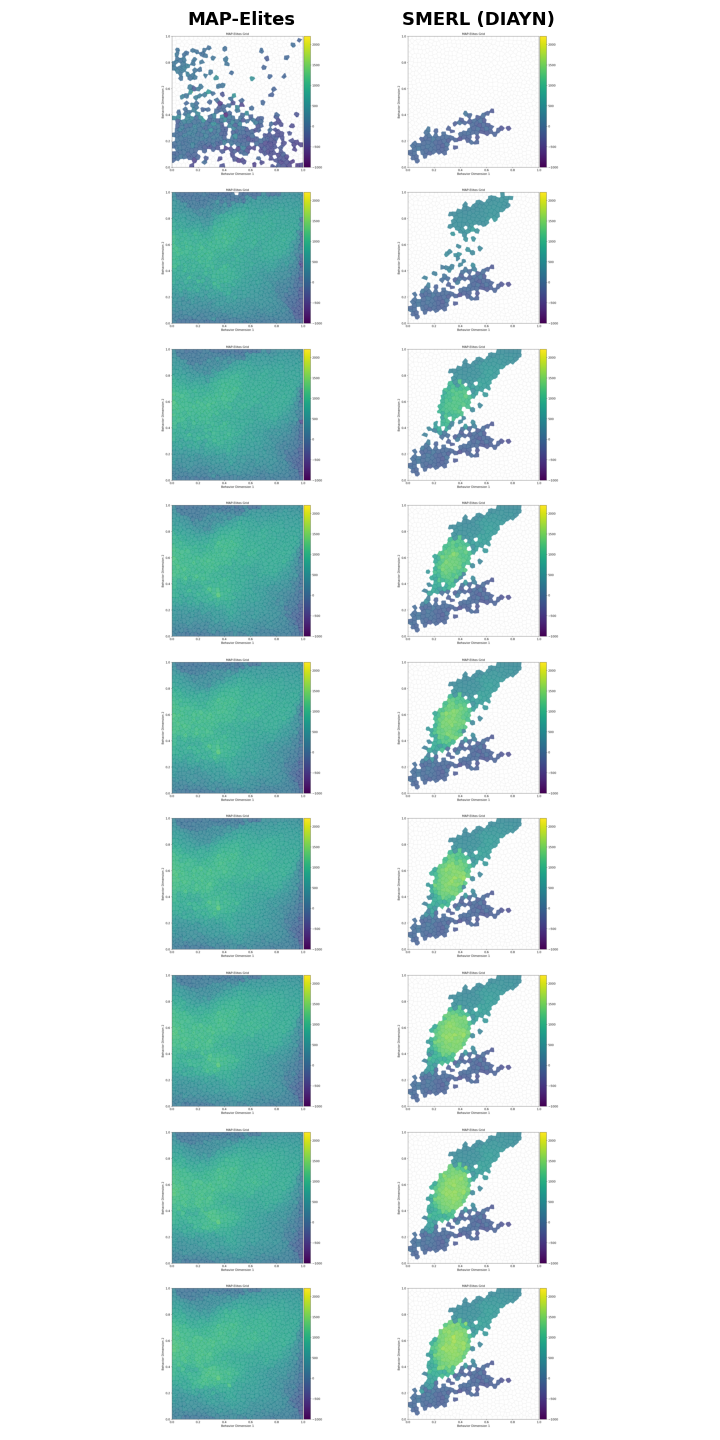}
         \caption{MAP-Elites and SMERL(DIAYN) on HalfCheetah-Uni.}
         \label{fig:grids_halfcheetah_uni}
     \end{subfigure}
     \begin{subfigure}{0.49\textwidth}
        \captionsetup{width=.85\linewidth}
         \hspace{-29pt}
         \includegraphics[height=0.8\textheight]{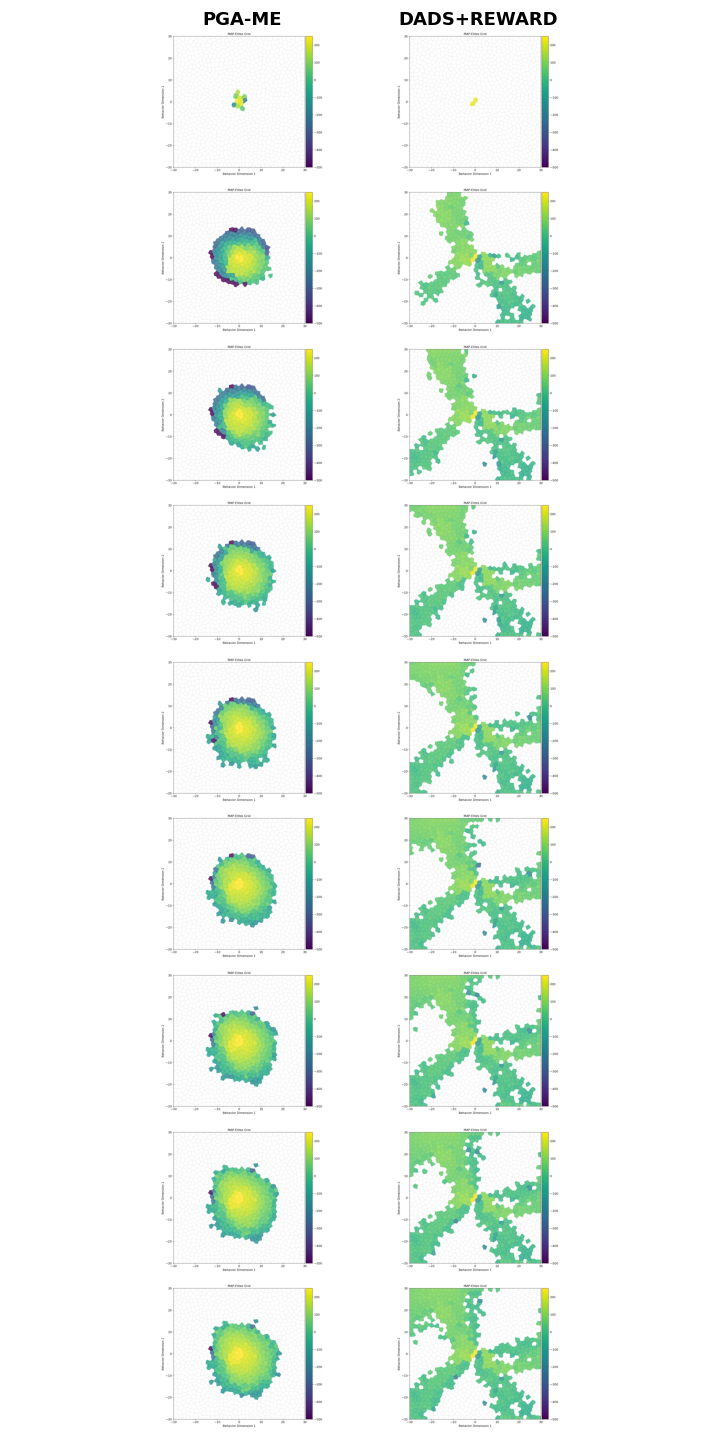}
         \caption{PGA-MAP-Elites and DADS+REWARD on Ant-Omni.}
         \label{fig:grids_ant_omni}
     \end{subfigure}
     \caption{Intermediate repertoires. Visualizations were generated at regular intervals from the start of training (top) to halfway through the training phase (i.e. at the 1h mark) (bottom).}
     \label{fig:three graphs}
\end{figure}

We provide further comparison in the way the repertoire are filled by the different methods. \figureautorefname~\ref{fig:grids_halfcheetah_uni} highlights the capability of \me to fill the whole descriptor space, but it struggles getting excellent policies whereas \diaynsmerl lacks coverage and can only evolve a small population of elite policies. On \antomni, the reward signal corresponds to the robot control cost, which hence gives incentives for minimal movement.

Nevertheless, we can see on \figureautorefname~\ref{fig:grids_ant_omni} how \pgame and \dadssum eventually explore the environment. \pgame uses the divergent search of genetic mutations to gradually find exploratory behaviors and although it fails to go far from the origin, it is able to build a dense set of behaviors. On the other hand, \dadssum uses the diversity rewards to discover far region of the behavior space through directed skills. Both methods have efficient strategies to keep exploring even if the environment rewards discourage them to do so.

While we have extended the concept of repertoires to skill-discovery methods in Section \ref{sec:diversity} to be able to compare all methods with the same metrics, an alternate way to assess the diversity achieved by the skill-discovery methods is to directly visualize the trajectories of the learned skills. On Figure \ref{fig:diayn_dads_skills}, we show the trajectories of the 5 skills learned by \diaynsum and \dadssum at the end of the training phase. On this particular seed, note that \diaynsum has been able to learn 5 skills that are significantly distinct from one another, with one of them able to sidestep the trap and run far along the x-axis direction. Conversely, \dadssum could hardly learn 5 distinct skills but managed to find two significantly distinct skills that go around the trap on opposite sides. Nevertheless, \figureautorefname~\ref{fig:dads_sum_skill_ant_omni} shows how \dadssum is able to learn skills going in 5 perfectly distinct directions in \antomni.

\section{Additional observations for the training phase}
\label{app:additional_obs_training}

It is worth mentioning that all skill-discovery \rl methods achieve higher fitnesses than the \qd methods on \antmaze even though they fail to fully explore the behavior descriptor space, as seen on Figure \ref{fig:repertoires}. Amusingly, watching the policies trained by these methods interact with the simulator reveals that they all have found a way to jump above a wall of the maze, thereby directly reaching the target position without having to solve the maze. Although the task was not designed to be solved this way, it is interesting to see that some algorithms were able to find an original way to solve the task. Running a pure optimization \rl algorithm with no mechanism for diversity does not find this way to jump above the maze.

It is interesting to see that Skill Discovery \rl methods get best performance on 4 tasks collectively, but only get one best fitness each. No method has a consistent best performance on the tasks.

For the \unidir tasks, we observe that the skill-discovery methods have been able to illuminate an important portion of the space even though they are not explicitly looking for diversity w.r.t. the behavior descriptors. However, note that they have not been able to find solutions at the edges of the repertoire, whose cells can only be occupied by solutions consisting in walking on only one feet or in an asymmetric manner which we expect to be more robust to perturbations.

Lastly, note that \me outperforms all other methods w.r.t. the \qd score for 3 out of the 7 tasks while achieving similar coverage than \pgame, which means that it has been able to find high-performing solutions. This is a surprising fact in itself because this method does not leverage back-propagation through the neural policies so it is expected to be less efficient than the other methods. This is a testament to the efficiency of the vectorized \jax implementation that enables \me to iterate over large populations of policies very quickly.

\section{Additional results for the adaptation experiments}

\begin{figure}[h!]
    \centering
    \includegraphics[width=1.\textwidth]{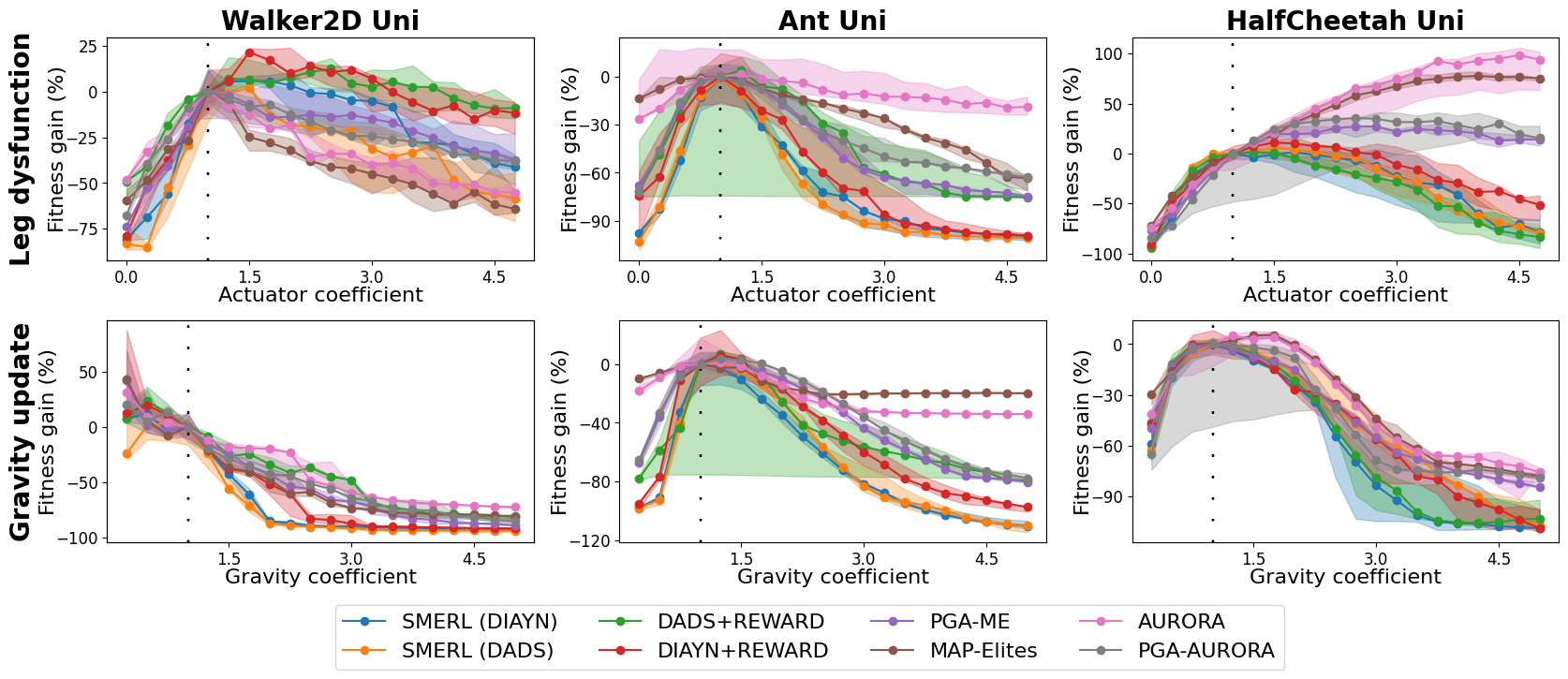}
    \caption{Relative change in fitness, referred to as \emph{fitness gain}, on the adaptation tasks. Comparing the algorithms' performances on modified environments to their own performances on the nominal environments offers another perspective on the resilience capabilities of the methods under study.}
    \label{fig:adaptation_fitness_loss}
\end{figure}

In this section, we provide additional results for the environment adaptation experiments described in Section \ref{sec:downstream}. Specifically, we plot the relative change in fitness, referred to as \emph{fitness gain}, compared to the nominal scenario (where the environment is unchanged) on Figure \ref{fig:adaptation_fitness_loss}. This is to be compared with the raw fitness reported on Figure \ref{fig:adapt_exps} in the main body of the paper. Observe on Figure \ref{fig:adaptation_fitness_loss} that the fitness gain is always $0$ when the gravity (resp. actuator) coefficient is $1$, this is by definition because the environment is unmodified in this case. We expect the fitness gain to be negative in most cases, because agents should perform worse on environments they were never trained on. Nevertheless, we can see that, in some cases (small gravity values or highly sensitive actuators), some methods find policies that can make the best of these changes and actually perform better than on the environment they were trained on.

Figure \ref{fig:adaptation_fitness_loss} also provides insights into the sensitivity of the various methods to environmental changes. \me, while lagging behind in terms of raw performance compared to other methods, is particularly resilient in the face of environment perturbations. For instance, for the \antuni task, the performance of \me drops by at most 20\% on the full range of gravity values used for evaluation, whereas the fitnesses of most methods drop by at least 70\% in the most extreme conditions on both ends of the spectrum.

\section{Comparing supervised skill-discovery methods}
This paper uses four methods inspired from the skill-discovery literature: \diaynsum and \dadssum sum their diversity terms with the extrinsic reward from the environment, whereas \diaynsmerl and \dadssmerl optimize the diversity reward only if the skill is already close to be optimal w.r.t. the extrinsic reward. We can draw three conclusions about the relative merits of each approach from the results obtained in this paper.

First, during the training phase, we can see, on Table~\ref{qd-score-comparison} as well as on \figureautorefname~\ref{fig:qd_metrics_time}, that \smerl methods outperform "sum"-based methods on most environments w.r.t. fitness, with the notable exception of \anttrap where \dadssum outperforms any other method by a significant margin. On the contrary, we can see that "sum"-based methods achieved higher \qd scores than the \smerl methods on this benchmark. These results seem coherent with the fact that \smerl prioritizes optimizing for quality over optimizing for diversity. Nevertheless, this choice also impacts the exploration capabilities of \smerl methods which, in turn, affects their fitness performances on some environments that require a good amount of exploration to be solved, such as \anttrap. \smerl methods fail to find a way around the trap in  \anttrap whereas the "sum"-based methods eventually find one. Even if this improved exploration capability does not systematically lead to improved fitness in all environments, it helps "sum"-based methods attain higher \qd scores.

Second, in the adaptation experiments, \smerl methods outperform "sum"-based methods as long as the perturbations remain small but do not necessarily perform better in extreme scenarios. On \antuni for instance, \diaynsmerl outperforms \diaynsum by a significant margin (more than 20\% of maximum fitness, see \figureautorefname~\ref{fig:adapt_exps}) when the actuator coefficient is perturbed by less than 25\%, but for perturbations greater than 60\%, \diaynsum eventually outperform (for smaller values) or draw (for bigger values) \diaynsmerl. We can observe similar tendencies on the gravity update experiment.

Third, "sum"-based methods seem to suffer from higher variances than \smerl methods. The performance of \dadssum on \antuni (see \figureautorefname~\ref{fig:adapt_exps}) is an example of this tendency on the adaptation tasks. We also observed a significantly higher variance of \dadssum and \diaynsum on several environments during the training phase, for all three metrics reported on \figureautorefname~\ref{fig:qd_metrics_time}.

Although the benchmark carried out in this paper confirms the benefits of \smerl \citep{kumar2020one} for adaptation tasks, it also shows that there is room for further improvements for these methods, particularly in settings where the task or the environment is significantly perturbed after training.

\section{Hyperparameters sensitivity}
\label{app:hyperparams_sensitivity}

In this section, we detail the hyperparameter-sensitivity experiment presented in Section \ref{sec:diversity} whose results are shown on \figureautorefname~\ref{fig:hyperparam_se}. Recall that this experiment was motivated by the fact that \dadssum achieves impressive results on the \antomni task, covering a significant portion of the behavioral descriptor space, and outperforms all \qd methods - let alone other skill-discovery methods, see \figureautorefname~\ref{fig:repertoires}. Although it is not entirely surprising that \dads-related methods perform better than \diayn-inspired methods, we were surprised to see such a discrepancy in the results and we suspected that the results would be rather sensitive to the exact choice of hyperparameters. For this reason, as described in Section~\ref{sec:diversity}, we decided to run a study on the hyperparameter sensitivy of \dadssum on \antomni, comparing it to \me, the second most performing method in terms of \qd score. To do so, we selected, for both methods, hyperparameters that impact the updates of the policies and we ran a grid search on them to compare the distributions of \qd scores. For \dadssum, we selected the hyperparameters that impact the reward signal, as they are key to trade off diversity for fitness, and used three different values for each hyperparameter, within an acceptable range, based on the literature. We picked the diversity reward scaling value (referred to as $\beta$ in Section~\ref{sec:hyperparams}) in the set $\{0.1, 1, 10\}$. We picked the entropy temperature coefficient $\alpha$ in the set $\{0.1, 0.5, 1\}$, as it directly impacts the randomness of skills. For \me, the distribution of evolved policies is highly dependant on the exact definition of the genetic variations used as subroutines. Hence, we decided to run a grid search on the iso (resp. line) sigma values of the Iso+LineDD variation \citep{vassiliades2018discovering} used in \me. Iso sigma parameter were selected in the set $\{0.001, 0.01, 0.1\}$ and line sigma were selected in the set $\{0.01, 0.1, 1.0\}$. For each algorithm, this yields 9 different possibilities. Each possibility is evaluated on 5 different seeds. The results are reported on \figureautorefname~\ref{fig:hyperparam_se}.

\begin{figure}[h!]
    \centering
    \includegraphics[width=1.\textwidth]{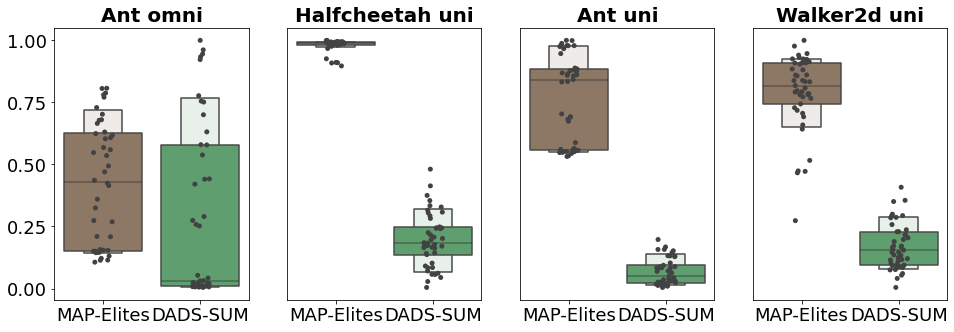}
    \caption{\qd scores obtained when averaging over a grid of hyperparameters on \antomni, \hcuni, \antuni and \walkeruni. \qd scores are normalized with the maximal values obtained on each task. Each point represent one set of hyper-parameters. Main axes represent the 0.25 and 0.75 quantiles, the horizontal bar represents the median and the light color boxes represent the 0.125 and 0.875 quantiles.}
    \label{fig:hyperparams_sensitivity_full}
\end{figure}

In order to strengthen this study, we decided to run the same experiments on the three \unidir tasks as well: \antuni, \walkeruni and \hcuni. We present the results on all the four environment on \figureautorefname~\ref{fig:hyperparams_sensitivity_full}.

On \hcuni, we can see that \me is being very consistent over the set of hyperparameters, with a high median and a very low variance, where \dadssum has a low median and a rather high variance. On the \walkeruni, both methods have a similar variance over the hyperparameters but \me is consistently performing better. On \antuni, \me has a significantly higher variance but it's worse case still outperforms \dadssum's best \qd score on the task. Those results shows that \me is easy to use off-the-shelf as it tends to have a smaller variance and as its worse case performance are usually decent. This makes this family algorithm very convenient for industrial use.

\section{Unsupervised behavior discovery in Quality Diversity methods}

The two Quality Diversity methods presented in the paper, \me and \pgame, are using a predefined behavior descriptor, which suppose some prior knowledge about the task. Skill-discovery RL methods often use a prior too (for instance, the xy position of the center of gravity, which is what we did in this study as well), they can be applied without it if needed.

Nevertheless, recent works from the \qd community introduce methods to learn the behavior descriptor in an unsupervised manner \citep{cully2019aurora, grillotti2022aurora}. One of these methods, called \aurora, stores the trajectories of the policies of its population, and use a Variational Autoencoder (\vae) \citep{Kingma2014AutoEncodingVB} to encode and try to reconstruct the trajectory. The encoder is then used to encode any new policy's trajectory into the latent space of the \vae.

Several other recent works build on this approach and use other information to learn automatically the behavior descriptor from data. \ruda \citep{grillotti2022ruda} builds on \aurora but bias the behavior descriptor towards behaviors that are helpful in an adaptation tasks. A similar approach is explored in the multi-agent setting by \citet{dixit2022diversifying}, to evolve population of agents where the diversity improves the collaboration in a common task.

In order to validate the interest of those methods, we decided to implement \aurora and to compare its performance with \me in several training and adaptation tasks, to see the potential loss of performance induced by this autonomous way to learn the behavior descriptor. We do not claim any exhaustive study of this method and let this for future work, but want to mention their existence and give an idea of its performance on some of our tasks.

The authors of \aurora were able to provide us a \jax implementation that worked on the \ant robots. To better assess the interest of those methods, we chose a task were no prior was given: \antuni. The data given to the \vae is made of trajectories in the environment, with the full observed state. The proportion of contact time is not given to the algorithm. To the best of our knowledge, \aurora had never been trained or used for adaptation on a \unidir task.

We decided to go further and use the unsupervised mechanism of \aurora to handle the diversity of the population of \pgame. This variant, that we refer to as \pgaaurora, is the first attempt to use gradient mechanism inside \aurora.

The results of \aurora and \pgaaurora are compared to all other methods is reported in \figureautorefname~\ref{fig:adapt_exps}.

Our results show that the unsupervised behavior discovery method introduced by \aurora resulted in very small performance loss compared to \me and that combining it with \pgame even show improved adaptation ability in the whole adaptation range on \antuni with the gravity update and 80\% of the adaptation range on the leg dysfunction task.

\section{Impact of the prior and of the environment reward signal on skill discovery methods}

\paragraph{Impact of the prior} The impact of prior in skill discovery methods has already been studied in the literature, for instance in the paper that introduced \dads \citep{sharma2019dynamics}. Nevertheless, we wanted to illustrate this effect on an example taken from one of the tasks of our study, \antomni.

Our setting is different from the setting used in \citet{sharma2019dynamics} because the algorithms are modified to take into account the reward of the environment. In this task, the reward corresponds to the opposite of the control cost.

\begin{wrapfigure}{r}{0.49\textwidth}
    \centering
    \includegraphics[width=.9\linewidth]{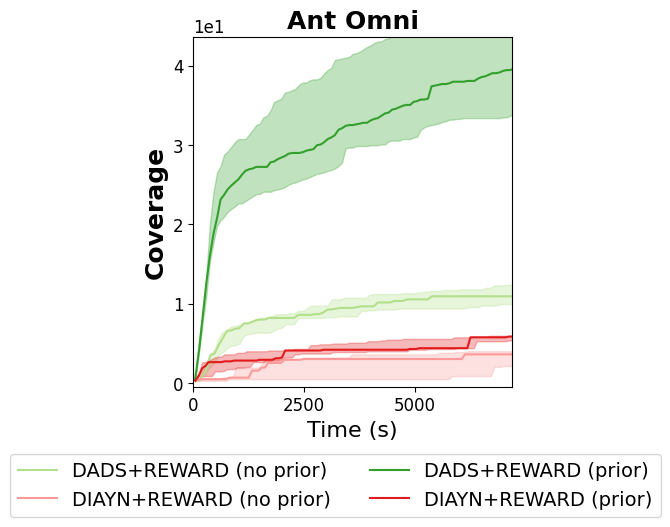}
    \caption{Effect of the prior on skill discovery methods in \antomni. Evolution of the coverage along time. The given prior is the $(x, y)$ position of the ant's center of gravity.}
    \label{fig:prior_impact_coverage}
\end{wrapfigure}

To illustrate the effect of the prior, we took the skill discovery methods that performed the best on \antomni with the prior: \diaynsum and \dadssum. We removed the prior, and trained the methods again on five seeds. Hence, instead of searching for diversity in the $(x, y)$-space, skills are looking for diversity in the entire state space of the \ant robot.

\begin{figure}[h!]
    \centering
    \begin{subfigure}{.22\linewidth}
        \centering
        \includegraphics[width=1\linewidth]{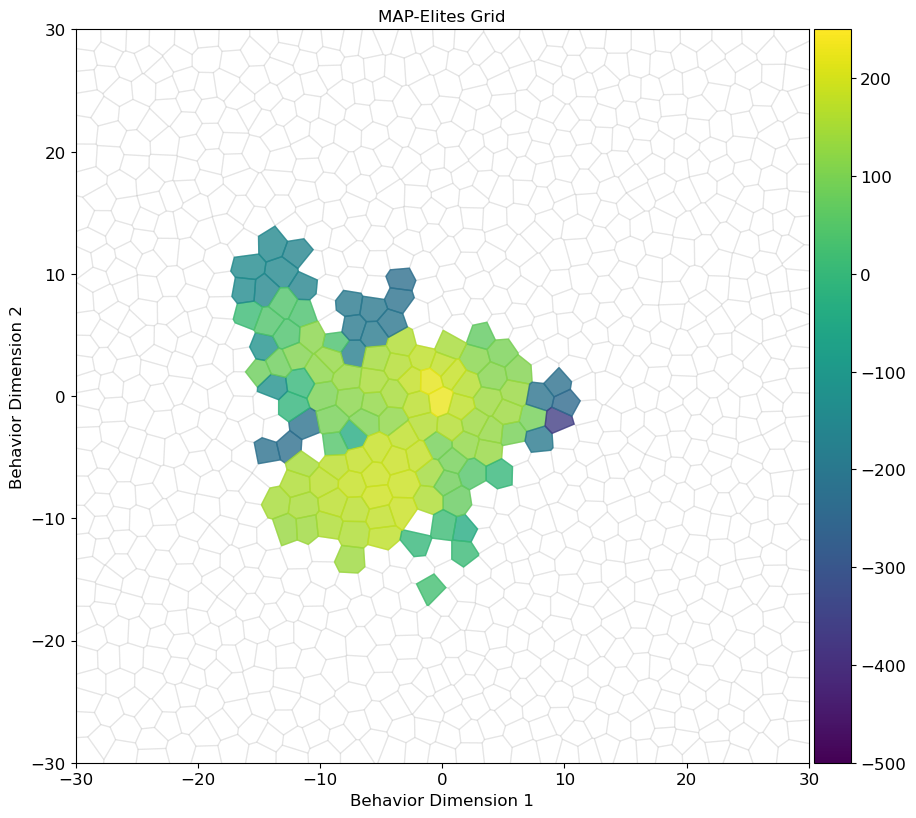}
        \caption{w/o prior}
    \end{subfigure}%
    \begin{subfigure}{.22\linewidth}
        \centering
        \includegraphics[width=1\linewidth]{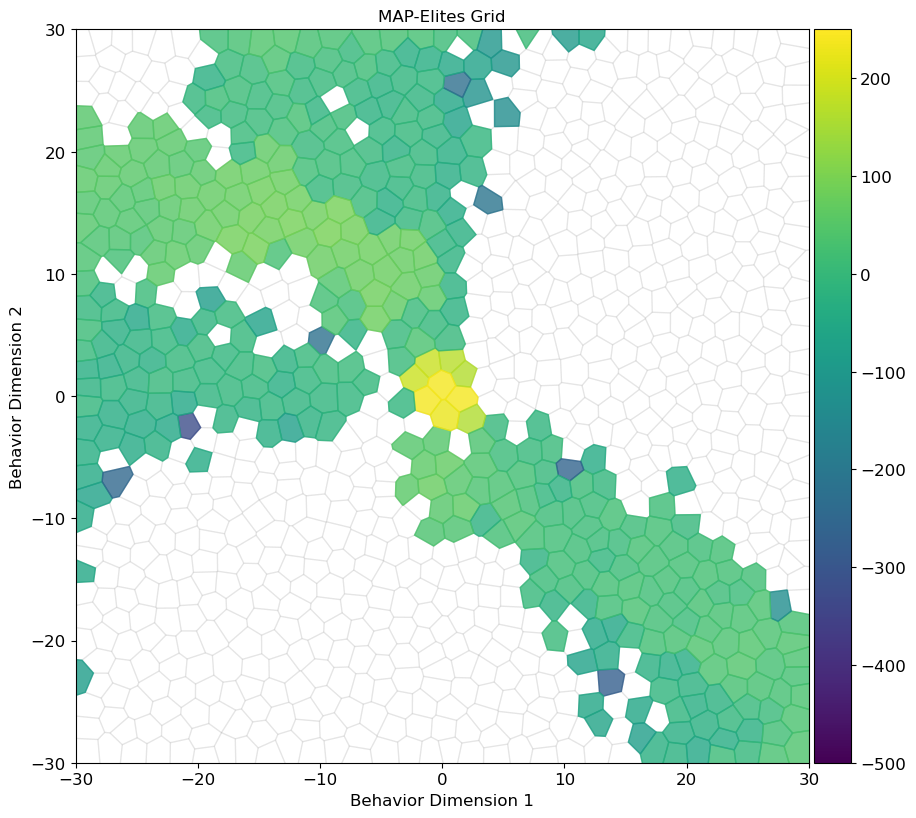}
        \caption{w/ prior}
    \end{subfigure}
    \hfill
    \begin{subfigure}{.22\linewidth}
        \centering
        \includegraphics[width=1\linewidth]{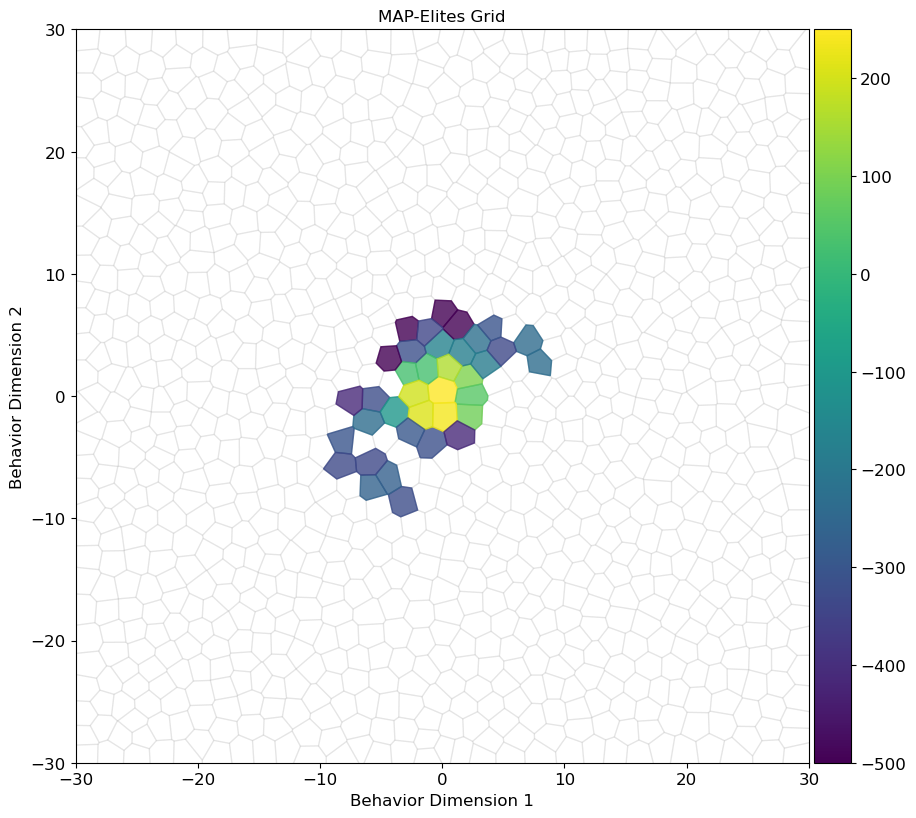}
        \caption{w/o prior}
    \end{subfigure}%
    \begin{subfigure}{.22\linewidth}
        \centering
        \includegraphics[width=1\linewidth]{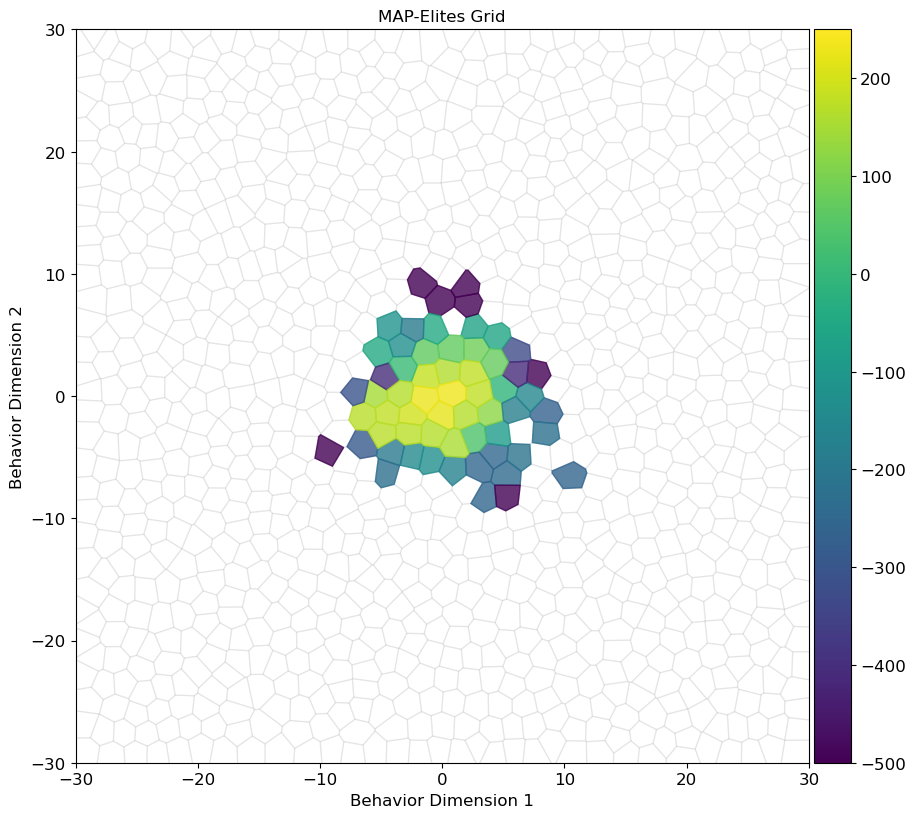}
        \caption{w/ prior}
    \end{subfigure}
    \hfill
    \begin{subfigure}{.22\linewidth}
        \centering
        \includegraphics[width=1\linewidth]{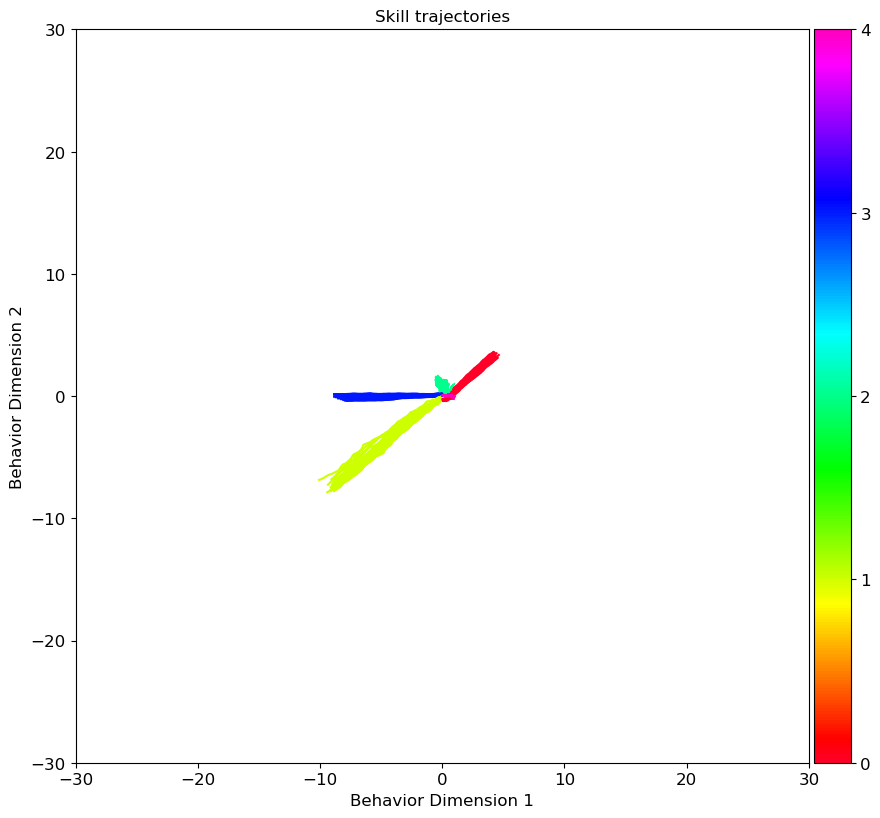}
        \caption{w/o prior}
    \end{subfigure}%
    \begin{subfigure}{.22\linewidth}
        \centering
        \includegraphics[width=1\linewidth]{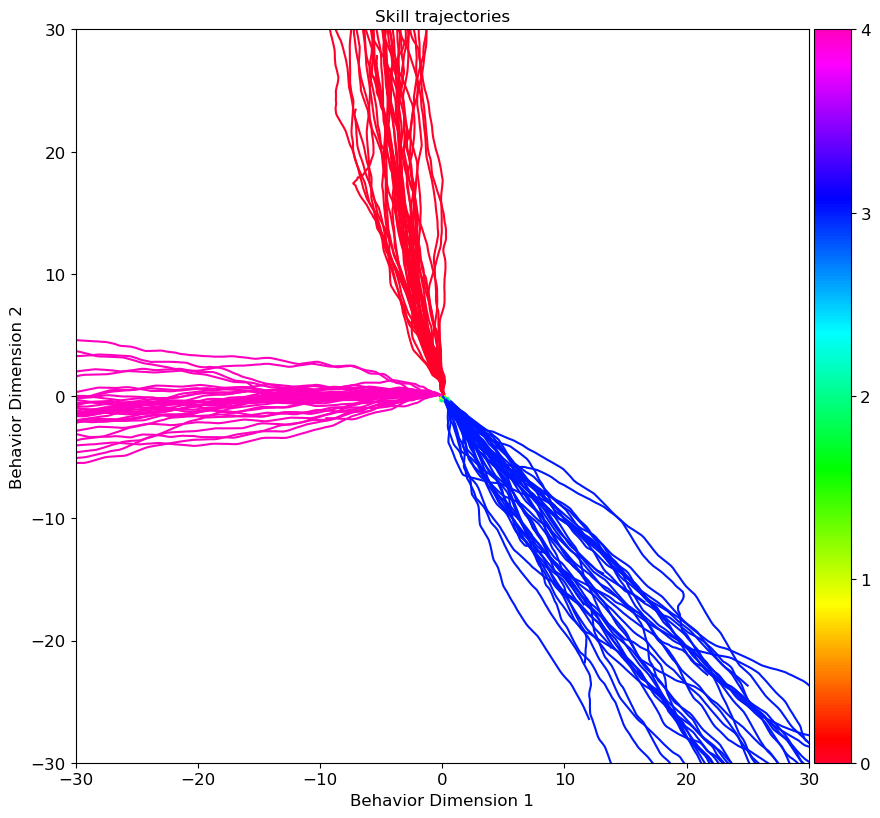}
        \caption{w/ prior}
    \end{subfigure}
    \hfill
    \begin{subfigure}{.22\linewidth}
        \centering
        \includegraphics[width=1\linewidth]{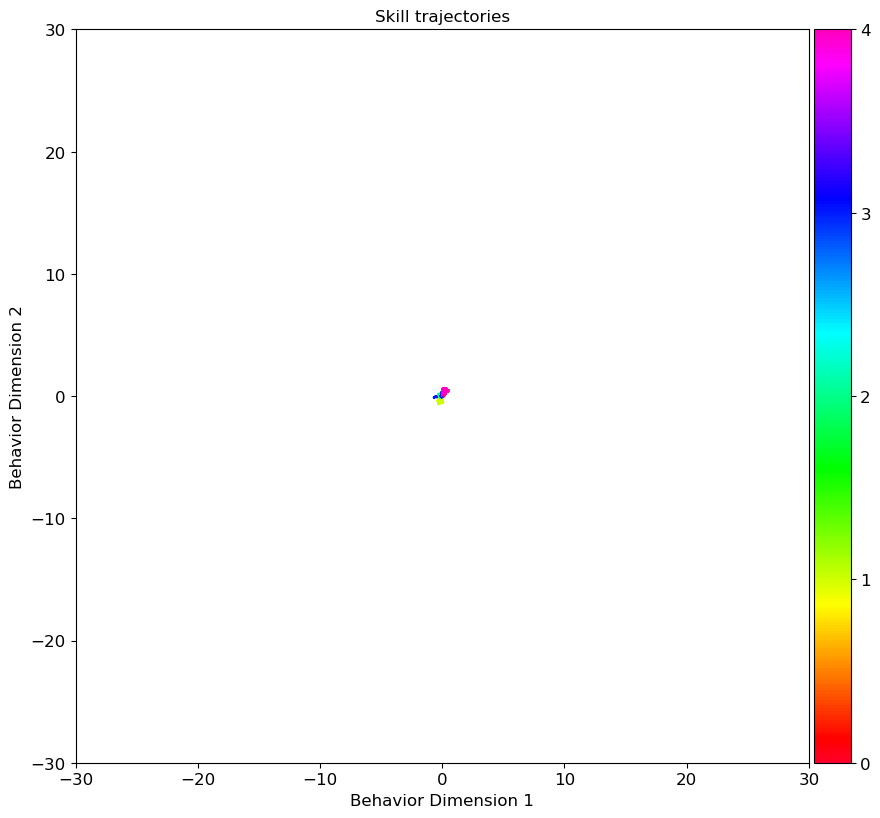}
        \caption{w/o prior}
    \end{subfigure}%
    \begin{subfigure}{.22\linewidth}
        \centering
        \includegraphics[width=1\linewidth]{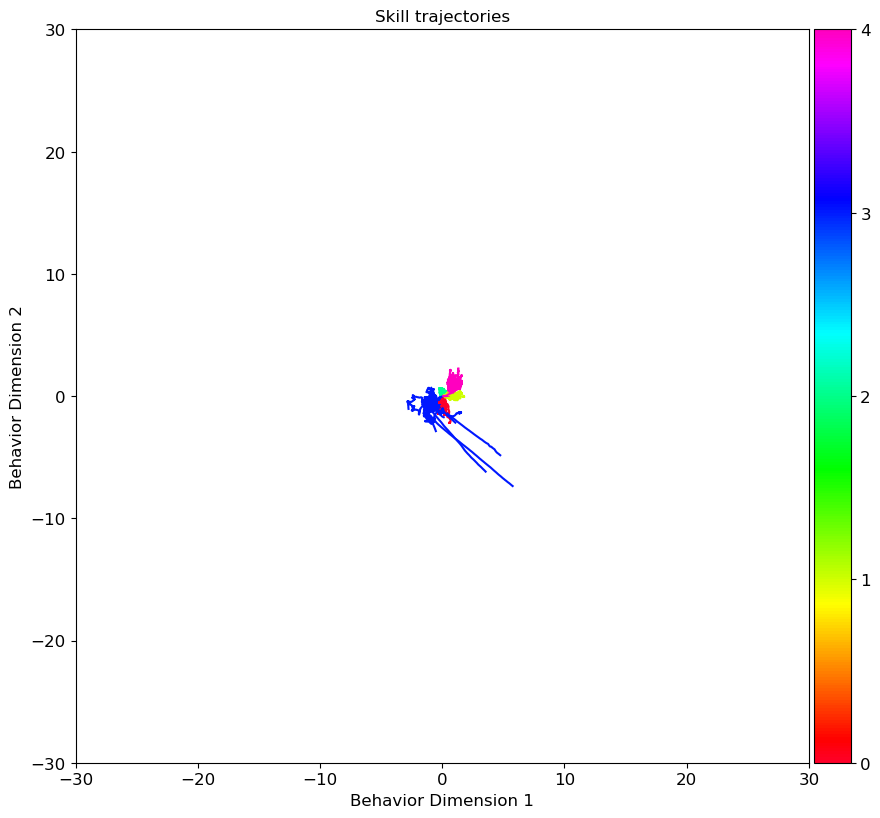}
        \caption{w/ prior}
    \end{subfigure}
    \caption{Repertoire and skills obtained by \dadssum (left, (a-b-e-f)) and \diaynsum (right, (c-d-g-h)) in \antomni with or without the prior. Using a prior narrows the space where the method is trying to find diversity. When no prior is given, the entire state space is considered. In \antomni, the prior is the $(x,y)$ position of the ant's center of gravity.}
    \label{fig:diayn_dads_prior_effect}
\end{figure}

On \figureautorefname~\ref{fig:prior_impact_coverage}, we report the evolution of the coverage of all methods along time. On \figureautorefname~\ref{fig:diayn_dads_prior_effect}, we show the final repertoire and skills of the median obtained for \diaynsum (resp. \dadssum) in terms of coverage.

The evolution of coverage on \figureautorefname~\ref{fig:prior_impact_coverage} shows that the final value is typically divided by four for \dadssum and decreased by 40\% for \diaynsum. The final repertoire and skills, \figureautorefname~\ref{fig:diayn_dads_prior_effect}, illustrate how consistently the methods are able to navigate in the environment with or without the prior.
The impact of the prior would be smaller in environments were the extrinsic reward can help to explore but it would still have significant impact.

\begin{table}[h!]
  \caption{Maximum fitness and \qd score at the end of the training phase (median over 5 seeds).}
  \centering
  \resizebox{.98\textwidth}{!}{
      \centering
    
      \begin{tabular}{lcccccc}
        \toprule
             & \anttrap & \antmaze & \antuni & \antomni & \walkeruni & \hcuni \\
        \midrule
            \multicolumn{7}{c}{\textbf{Maximum fitness}}\\
        \midrule
        \diaynsmerl & 4.49 ($\pm$0.01)\text{e}2 & -7.88 ($\pm$0.41)\text{e}3  & \textbf{1.71 (}$\pm$\textbf{0.09)e3} &\textbf{2.49 ($\pm$0.00)e2} &\textbf{1.19 ($\pm$0.06)\text{e}3} &1.89 ($\pm$0.14)\text{e}3 \\
        \dadssmerl &4.48 ($\pm$0.02)\text{e}2 & \textbf{-7.62 ($\pm$0.43)e3} &1.61 ($\pm$0.18)\text{e}3 &\textbf{2.49 ($\pm$0.00)e2} & 1.17($\pm$0.14)\text{e}3 & 1.92 ($\pm$0.21)\text{e}3\\
        \dadssum & \textbf{1.28 ($\pm$0.41)e3} & -7.81 ($\pm$0.49)\text{e}3 & 1.24 ($\pm$0.52)\text{e}3 & 2.38 ($\pm$0.11)\text{e}2 & 6.30 ($\pm$0.48)e2 & 1.89 ($\pm$0.03)\text{e}3 \\
        \diaynsum & 9.42 ($\pm$3.43)\text{e}2 & -8.13 ($\pm$0.51)\text{e}3 & 1.34 ($\pm$0.21)\text{e}3 & 2.29 ($\pm$0.04)\text{e}2 & 1.00 ($\pm$0.13)\text{e}3 & \textbf{2.13 ($\pm$0.27)e3} \\
        \diayn & 2.46 ($\pm$0.09)\text{e}2 & -1.58 ($\pm$0.06)\text{e}4 & 2.19($\pm$0.70)e2 & 2.11 ($\pm$0.24)e2 & 3.42 ($\pm$0.20)\text{e}2 &4.63 ($\pm$0.61)\text{e}2 \\
        \dads & 2.46 ($\pm$3.84)\text{e}2 & \text{-9.16 ($\pm$1.29)e3} &2.18 ($\pm$0.72)\text{e}2 & \text{2.11 ($\pm$0.68)e2} & 3.50($\pm$0.74)\text{e}2 & 4.04 ($\pm$1.43)\text{e}2 \\
        \midrule
            \multicolumn{7}{c}{\textbf{QD score}}\\
        \midrule
        \diaynsmerl &1.02 ($\pm$0.17)\text{e}5 & 7.84 ($\pm$0.44)\text{e}5 & 6.20 ($\pm$1.38)\text{e}4 &1.95 ($\pm$1.70)\text{e}3 &3.01 ($\pm$0.20)\text{e}5 &7.92 ($\pm$2.23)\text{e}5\\
        \dadssmerl &9.88 ($\pm$0.56)\text{e}4 & 8.22 ($\pm$0.07)\text{e}5 & 0.83 ($\pm$1.17)\text{e}5 & 4.15 ($\pm$1.11)\text{e}3 & 2.69 ($\pm$0.38)\text{e}5 & 7.51 ($\pm$1.60)\text{e}5\\
        \dadssum & \textbf{2.42 ($\pm$0.80)\text{e}5} & 9.12 ($\pm$0.76)\text{e}5 & \textbf{2.88 ($\pm$1.07)\text{e}5} & 3.05 ($\pm$1.22)\text{e}5 & 3.89 ($\pm$0.41)\text{e}5 & 8.31 ($\pm$3.91)\text{e}5\\
        \diaynsum & 2.27 ($\pm$0.62)\text{e}5 & \textbf{9.18 ($\pm$0.16)\text{e}5} & 1.29 ($\pm$0.39)\text{e}5 & 4.00 ($\pm$0.39)\text{e}4 & \textbf{4.92 ($\pm$0.35)\text{e}5} & \textbf{1.68 ($\pm$0.18)\text{e}6} \\
        \diayn &5.80 ($\pm$1.74)\text{e}4 &9.35 ($\pm$2.83)\text{e}4 & 7.46 ($\pm$1.29)\text{e}4 & 4.84 ($\pm$4.50)\text{e}4 &1.02 ($\pm$0.20)\text{e}5 &4.00 ($\pm$0.50)\text{e}5 \\
        \dads & 1.37 ($\pm$0.14)\text{e}5 &6.99 ($\pm$2.32)\text{e}5 & 1.13 ($\pm$0.29)\text{e}5  & \textbf{3.50 ($\pm$0.56)\text{e}5}  & 2.67 ($\pm$0.62)\text{e}5 & 7.97 ($\pm$1.13)\text{e}5 \\
        \bottomrule
      \end{tabular}
}
  \label{tab:max_fitness_no_reward}
\end{table}

\paragraph{Impact of the environment reward signal} Throughout our study, we considered Skill Discovery methods extended to a supervised setting where they take into account the environment reward signal either by summation or with the \smerl method. We advocate that evaluating \dads and \diayn without this additional supervision would be unfair, since other methods explicitly optimize for the environment's reward. To validate this point, we train \dads and \diayn in a fully unsupervised setting and measure their performance. In \tableautorefname~\ref{tab:max_fitness_no_reward}, we report the maximum fitness and the QD-score obtained by these methods after 2-hours of training. Except on \antomni (where the environment reward is a penalty for moving), the best score is always attained by a supervised method, in terms of maximum fitness as well as QD-score.

\begin{figure}[h]
    \centering
    \includegraphics[width=\textwidth]{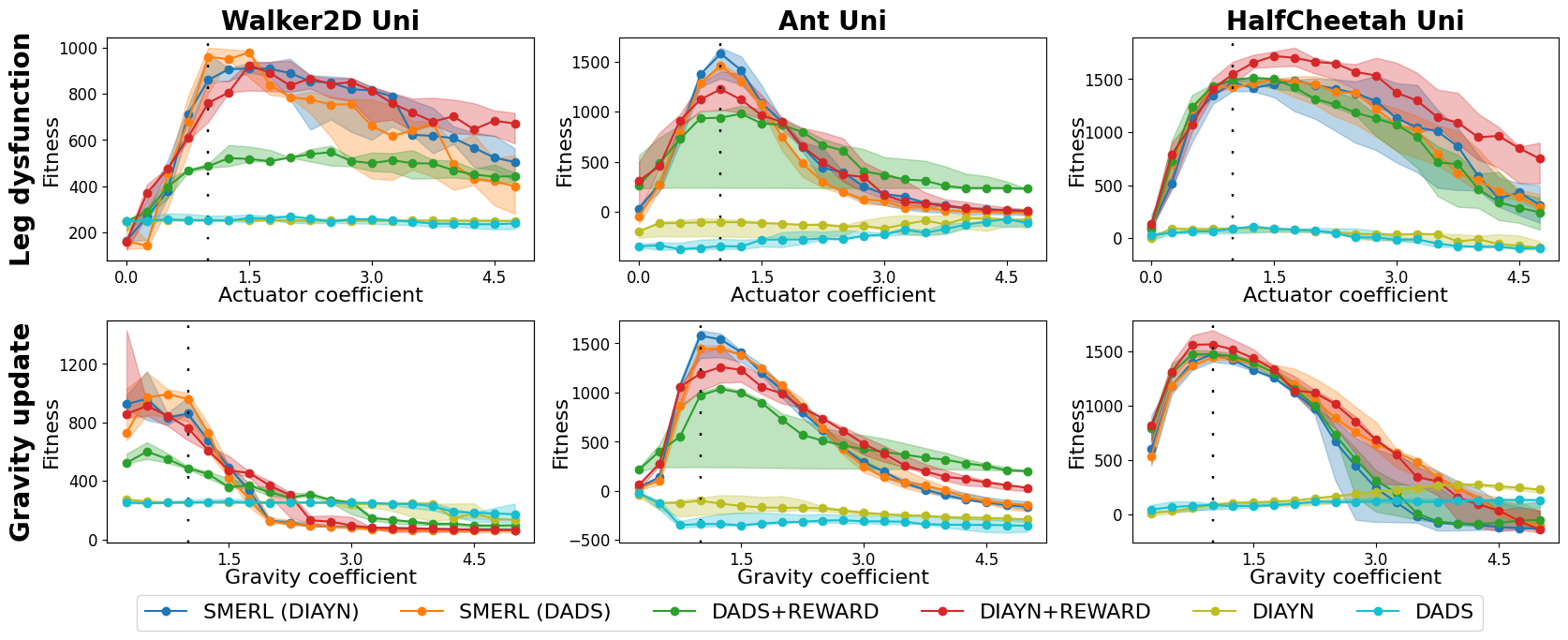}
    \caption{Effect of the environment reward signal on the algorithms \dads and \diayn on adaptation tasks. Results are the median and interquartile range over 5 random seeds.}
    \label{fig:reward_impact_adapt}
\end{figure}

To further emphasize the importance of the environment reward signal, we run adaptation experiments with \dads and \diayn and evaluate the robustness of the learned policies on perturbed environments. On \figureautorefname~\ref{fig:reward_impact_adapt} we report the adaptation results, and compare \dads and \diayn to their supervised counterparts \diaynsum, \diaynsmerl, \dadssum and \dadssmerl. It is striking how the performance of \diayn and \dads plummet without the access to the environment reward signal. It confirms the intuition that without any environement reward signal or finetuning, \dads and \diayn fail to consistently learn high-performing policies and to adapt in the face of a perturbed environment's dynamics.

\section{Additional statistical analysis of the results}

In this section, we provide additional data concerning the results in Table~\ref{tab:max_fitness},  Figure~\ref{fig:adapt_exps} and Figure~\ref{fig:hurdles}.

\begin{figure}[h!]
    
    \begin{subfigure}{1.0\linewidth}
    \centering
    \includegraphics[width=0.95\textwidth]{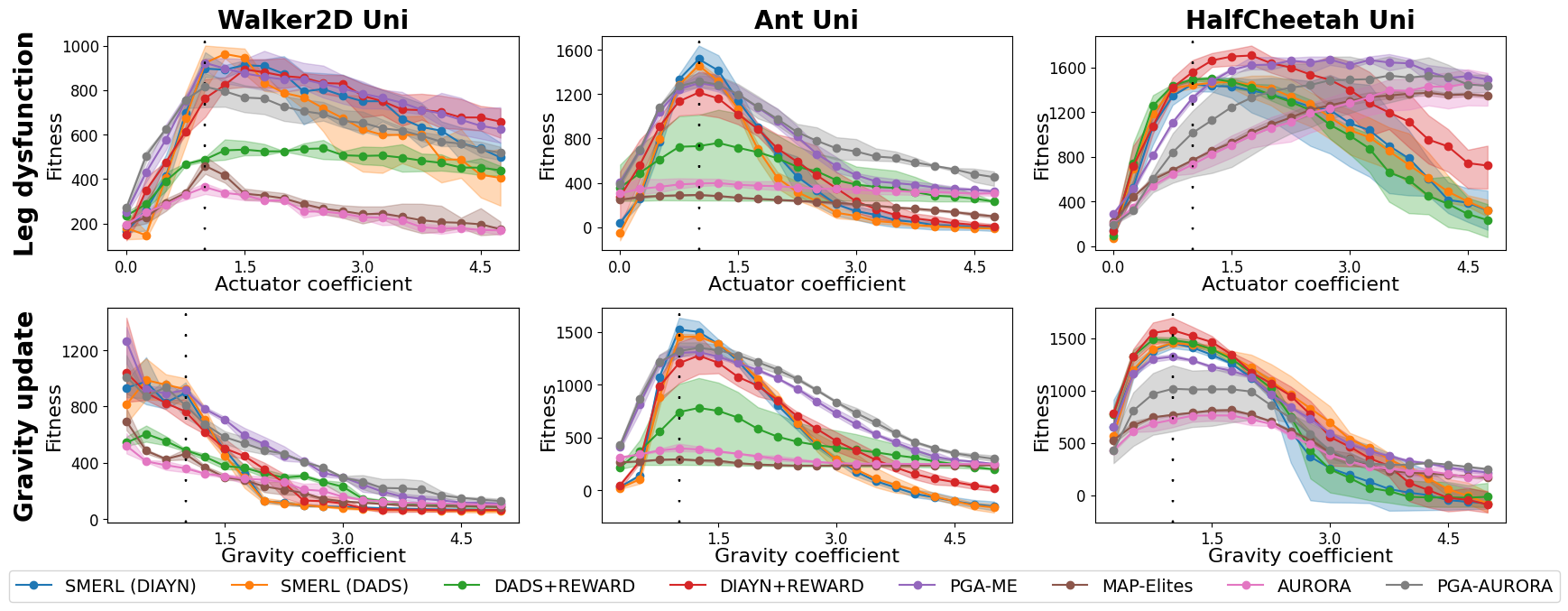}
    \end{subfigure}%
    
    \centering
    \begin{subfigure}{.45\linewidth}
        \centering
    \includegraphics[width=0.8\textwidth]{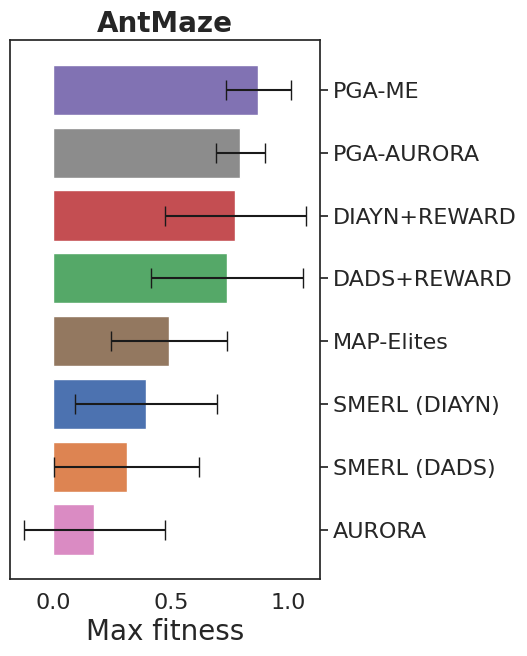}
    \end{subfigure}%
    \hfill
    \begin{subfigure}{.45\linewidth}
        \centering
    \includegraphics[width=1.\textwidth]{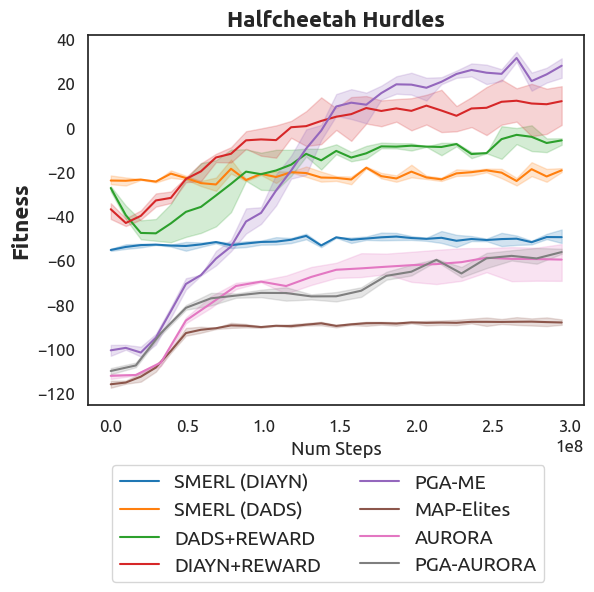}
    \end{subfigure}
    \caption{Maximum fitness of the methods when facing a perturbed environment (top) or a modified task (bottom-left) and fitness of the PPO meta-controller for various choices of skill-discovery methods used to generate primitives (bottom-right). IQM aggregation is used instead of median aggregation.}
    \label{fig:iqm_results}
\end{figure}

\paragraph{} In Figure~\ref{fig:iqm_results} we report the performance on various adaptation and hierarchical tasks, similarly to Figure~\ref{fig:adapt_exps} and Figure~\ref{fig:hurdles} but using IQM aggregation rather than median aggregation. We observe almost identical results for all benchmarks, setting aside \pgaaurora (resp. \diaynsum) that was ranked third (resp. second) for the \antmaze adaptation task and is now ranked second (resp. third).

\paragraph{} In Table~\ref{tab:max_fitness_iqm}, we include additional metrics corresponding to mean, median and IQM of the maximum fitness and the \qd score over the 5 seeds used in the training phase, whereas Table~\ref{tab:max_fitness} only includes median for readability purpose. Note that the best method for each environment is almost always the same irrespective of the exact statistic used.

\begin{table}[h]
  \caption{Final Max fitness and \qd score (training phase). Mean | median | IQM (5 seeds).}
  \centering
  \resizebox{.98\textwidth}{!}{
  \centering

  \begin{tabular}{lccccccc}
    \toprule
         & \anttrap & \antmaze & \ptmaze & \antuni & \antomni & \walkeruni & \hcuni \\
    \midrule
        \multicolumn{8}{c}{\textbf{Maximum fitness}} \\
    \midrule
\diaynsmerl & 4.49|4.49|4.49 \text{e}2 & -8.04|-7.88|-7.93 \text{e}3 & -5.33|-3.98|-5.30 \text{e}1 & \textbf{1.69}|\textbf{1.71}|\textbf{1.69} \text{e}3 & \textbf{2.49}|\textbf{2.49}|\textbf{2.49} \text{e}2 & 1.18|1.19|1.18 \text{e}3 & 1.87|1.89|1.91 \text{e}3\\
\dadssmerl & 4.48|4.48|4.48 \text{e}2 & \textbf{-7.56}|\textbf{-7.62}|\textbf{-7.66 }\text{e}3 & -5.76|-8.09|-6.15 \text{e}1 & 1.59|1.61|1.60 \text{e}3 & \textbf{2.49}|\textbf{2.49}|\textbf{2.49 }\text{e}2 & 1.10|1.17|1.13 \text{e}3 & 1.99|1.92|1.96 \text{e}3\\
\dadssum & \textbf{1.17}|\textbf{1.28}|\textbf{1.30} \text{e}3 & -7.65|-7.81|-7.78 \text{e}3 & -5.76|-8.09|-6.15 \text{e}1 & 1.13|1.24|1.25 \text{e}3 & 2.34|2.38|2.35 \text{e}2 & 6.12|6.30|6.12 \text{e}2 & 1.89|1.90|1.88 \text{e}3\\
\diaynsum & 7.92|9.42|8.14 \text{e}2 & -8.34|-8.13|-8.33 \text{e}3 & -5.52|-4.48|-5.65 \text{e}1 & 1.46|1.34|1.40 \text{e}3 & 2.29|2.29|2.29 \text{e}2 & 9.96|1.00|1.01 \text{e}3 & \textbf{2.05}|\textbf{2.12}|\textbf{2.05} \text{e}3\\
\pgame & 8.89|9.77|9.38 \text{e}2 & -9.15|-9.32|-9.24 \text{e}3 & \textbf{-2.26}|\textbf{-2.26}|\textbf{-2.25} \text{e}1 & 1.46|1.45|1.45 \text{e}3 & \textbf{2.49}|\textbf{2.49}|\textbf{2.49} \text{e}2 & \textbf{1.21}|\textbf{1.22}|\textbf{1.23} \text{e}3 & 1.74|1.75|1.76 \text{e}3\\

\pgaaurora & 4.70|4.73|4.68 \text{e}2 & -9.28|-9.27|-9.27 \text{e}3 & -4.84|-6.31|-5.10 \text{e}1 & 1.39|1.44|1.44 \text{e}3 &  \textbf{2.49}|\textbf{2.49}|\textbf{2.49} \text{e}2 & 0.99|1.05|1.01 \text{e}3 & 1.35|1.47|1.35 \text{e}3\\
\me & 3.77|3.81|3.77 \text{e}2 & -1.03|-1.01|-1.04 \text{e}4 & -2.47|-2.48|-2.47 \text{e}1 & 4.33|4.25|4.27 \text{e}2 & \textbf{2.49}|\textbf{2.49}|\textbf{2.49} \text{e}2 & 6.84|6.88|6.86 \text{e}2 & 1.27|1.29|1.27 \text{e}3\\

\aurora & 3.86|4.06|4.05 \text{e}2 & -1.11|-1.10|-1.09 \text{e}4 & -2.39|-2.35|-2.36 \text{e}1 & 5.36|5.05|5.26 \text{e}2 & \textbf{2.49}|\textbf{2.49}|\textbf{2.49} \text{e}2 & 5.45|5.71|5.45 \text{e}2 & 1.22|1.21|1.19 \text{e}3\\

    \midrule
        \multicolumn{8}{c}{\textbf{QD score}} \\
    \midrule
\diaynsmerl & 1.03|1.03|1.06 \text{e}5 & 7.69|7.84|7.71 \text{e}5 & 4.71|3.89|4.12 \text{e}4 & 6.77|6.19|6.66 \text{e}4 & 2.52|1.95|2.37 \text{e}3 & 3.07|3.01|3.06 \text{e}5 & 7.85|7.92|8.13 \text{e}5\\
\dadssmerl & 1.01|0.99|1.00 \text{e}5 & 8.60|8.22|8.38 \text{e}5 & 8.16|3.57|7.15 \text{e}3 & 1.25|8.32|7.96 \text{e}4 & 3.59|4.15|3.78 \text{e}3 & 2.78|2.69|2.67 \text{e}5 & 7.90|7.51|7.83 \text{e}5\\
\dadssum & 2.24|2.42|2.37 \text{e}5 & 8.51|8.22|8.38 \text{e}5 & 4.26|4.19|1.12 \text{e}4 & 2.55|2.88|2.81 \text{e}5 & 3.09|3.03|3.15 \text{e}5 & 3.85|3.89|3.83 \text{e}5 & 1.01|0.83|0.87 \text{e}6\\
\diaynsum & 2.31|2.27|2.25 \text{e}5 & 9.22|9.18|9.16 \text{e}5 & 3.70|4.04|3.55 \text{e}4 & 1.31|1.29|1.36 \text{e}5 & 3.92|4.00|3.97 \text{e}4 & 4.72|4.92|4.76 \text{e}5 & 1.77|1.68|1.73 \text{e}6\\
\pgame & \textbf{8.23}|\textbf{7.89}|\textbf{7.57} \text{e}5 & 2.72|2.74|2.80 \text{e}6 & 3.93|3.92|3.93 \text{e}5 & \textbf{9.19}|\textbf{9.19}|\textbf{9.18} \text{e}5 & 1.83|1.58|1.77 \text{e}5 & \textbf{8.03}|\textbf{8.23}|\textbf{8.11} \text{e}5 & 2.99|2.98|2.97 \text{e}6\\
\pgaaurora & 4.41|4.40|4.48 \text{e}5 & 2.62|2.59|2.58 \text{e}6 & 2.84|2.88|2.75 \text{e}5 & 7.34|7.33|7.37 \text{e}5 & 1.60|1.69|1.69 \text{e}5 & 5.00|5.10|5.07 \text{e}5 & 2.40|2.39|2.40 \text{e}6\\
\me & 4.74|4.53|4.63 \text{e}5 & \textbf{2.98}|\textbf{2.98}|\textbf{2.98} \text{e}6 & \textbf{4.23}|4.22|\textbf{4.22} \text{e}5 & 9.04|9.08|9.06 \text{e}5 & 3.00|3.05|3.11 \text{e}5 & 6.58|6.46|6.58 \text{e}5 & \textbf{3.05}|\textbf{3.05}|\textbf{3.05} \text{e}6\\
\aurora & 3.49|3.54|3.78 \text{e}5 & 1.85|2.18|1.95 \text{e}6 & 4.13|\textbf{4.58}|4.19 \text{e}5 & 6.15|5.82|6.01 \text{e}5 & \textbf{3.27}|\textbf{4.00}|\textbf{3.44} \text{e}5 & 4.46|4.59|4.46 \text{e}5 & 2.45|2.45|2.45 \text{e}6\\

    \bottomrule
  \end{tabular}
}
  \label{tab:max_fitness_iqm}
\end{table}

\clearpage

\section{Algorithms Pseudocodes}

We give the pseudocodes for all the algorithms used in our study. The main difference compared to the pseudocodes found in the original papers is that the termination of the algorithm is governed by a time condition. As a reminder, details of our implementations can also be found in Appendix~\ref{subsection:implem_algo}.

\makeatletter
\makeatother

\SetKwComment{Comment}{/* }{ */}

\SetArgSty{textnormal}

\begin{algorithm}
    \small
    \SetAlgoLined
    \DontPrintSemicolon
    \SetKwInput{KwInput}{Given}
    \KwInput{
    \begin{itemize}
        \item Sample size N, time allocation max\_time
        \item \me repertoire $\mathbb{M}$
        \item $N$ initial policies $\{\pi_{\theta_{i}}\}_{i=\{1,N\}}$\;
    \end{itemize}
    }


    \tcp{Main loop}
    $\textrm{current\_time} \leftarrow 0$\;
    \While{$\textrm{current\_time} < \textrm{max\_time}$}{
    \texttt{\\}
    
        \If{$\textrm{current\_time} > 0$}{
            \texttt{\\}
            \tcp{Sampling and mutation}
            Sample $N$ policies $\{\pi_{\theta_i}\}_{i = 1,N}$ from the repertoire $\mathbb{M}$\;
            Update the parameters $\{\theta_i\}_{i = 1,N}$ of the policies with genetic updates\;
            
        }
        \texttt{\\}
        \tcp{Evaluation and insertion in repertoire}
        Evaluate fitness and behavior descriptor of the new policies $\{\pi_{\widetilde{\theta_i}}\}_{i = 1,N}$\;
        Update current\_time\;
        Add the updated policies in the repertoire $\mathbb{M}$ when relevant\; 
    }

    \caption{MAP-Elites}
    \label{alg:ME}
\end{algorithm}

\makeatletter
\makeatother

\SetKwComment{Comment}{/* }{ */}

\SetArgSty{textnormal}

\begin{algorithm}
    \small
    \SetAlgoLined
    \DontPrintSemicolon
    \SetKwInput{KwInput}{Given}
    \KwInput{
    \begin{itemize}
        \item Sample size N, time allocation max\_time
        \item \me repertoire $\mathbb{M}$, Replay Buffer $\mathbb{B}$
        \item $N$ initial policies $\{\pi_{\theta_{i}}\}_{i=\{1,N\}}$ and
    a critic $Q_v$\;
    \end{itemize}
    }

    \tcp{Main loop}
    $\textrm{current\_time} \leftarrow 0$\;
    \While{$\textrm{current\_time} < \textrm{max\_time}$}{
    \texttt{\\}
    
        \If{$\textrm{current\_time} > 0$}{
            \texttt{\\}
            \tcp{Sampling and mutation}
            Sample N policies $\{\pi_{\theta_i}\}_{i = 1,N}$ in repertoire $\mathbb{M}$\;
            Sample batches of transitions in replay buffer $\mathbb{B}$\;
            Update half the policies using the critic $Q_v$\;
            Update the other half with genetic updates\;
            
            \texttt{\\}
            \tcp{Train the critic according to TD3 rule}
            Update the critic $Q_v$\;

        }
        \texttt{\\}
        \tcp{Evaluation and insertion in repertoire}
        Evaluate the new policies and store collected transitions in buffer $\mathbb{B}$\;
        Update current\_time\;
        Add the new policies in the repertoire $\mathbb{M}$\ when relevant; 
    }

    \caption{PGA-MAP-Elites}
    \label{alg:PGAME}
\end{algorithm}

\makeatletter
\makeatother

\SetKwComment{Comment}{/* }{ */}

\SetArgSty{textnormal}

\begin{algorithm}
    \small
    \SetAlgoLined
    \DontPrintSemicolon
    \SetKwInput{KwInput}{Given}
    \KwInput{
    \begin{itemize}
        \item Sample size $N$, time allocation max\_time
        \item Behavior extraction model $\xi_\theta$, training schedule train\_times
        \item \me repertoire $\mathbb{M}$, $N$ initial policies $\{\pi_{\theta_{i}}\}_{i=\{1,N\}}$\;
    \end{itemize}
    }
    \tcp{Main loop}
    $\textrm{current\_time} \leftarrow 0$\;
    \While{$\textrm{current\_time} < \textrm{max\_time}$}{
    \texttt{\\}
    
        \If{$\textrm{current\_time} > 0$}{
            \texttt{\\}
            \tcp{Sampling and mutation}
            Sample $N$ policies $\{\pi_{\theta_i}\}_{i = 1,N}$ from the repertoire $\mathbb{M}$\;
            Update the parameters $\{\theta_i\}_{i = 1,N}$ of the policies with genetic updates\;
            
        }
        \tcp{Evaluation and insertion in repertoire}
        Evaluate fitness and behavior descriptor of the new policies $\{\pi_{\widetilde{\theta_i}}\}_{i = 1,N}$\;
        Update current\_time\;
        Add the updated policies in the repertoire $\mathbb{M}$ when relevant\;
        
        \texttt{\\}
        \tcp{Training the behavior extraction model}
        \If{$\textrm{current\_time in train\_times}$}{
            Update $\xi_\theta$ to reconstruct trajectories of policies in $\mathbb{M}$\;
            Re-evaluate fitness and behavior descriptor of the stored policies\;
            Update repertoire $\mathbb{M}$ with new behavior descriptors\;
        }
    }

    \caption{AURORA}
    \label{alg:aurora}
\end{algorithm}

\makeatletter
\makeatother

\SetKwComment{Comment}{/* }{ */}

\SetArgSty{textnormal}

\begin{algorithm}
    \small
    \SetAlgoLined
    \DontPrintSemicolon
    \SetKwInput{KwInput}{Given}
    \KwInput{
    \begin{itemize}
        \item Sample size $N$, time allocation max\_time
        \item Behavior extraction model $\xi_\theta$, training schedule train\_times
        \item \me repertoire $\mathbb{M}$, $N$ initial policies $\{\pi_{\theta_{i}}\}_{i=\{1,N\}}$
        \item Replay Buffer $\mathbb{B}$ and
    critic $Q_v$\;
    \end{itemize}
    }

    \tcp{Main loop}
    $\textrm{current\_time} \leftarrow 0$\;
    \While{$\textrm{current\_time} < \textrm{max\_time}$}{
    \texttt{\\}
    
        \If{$\textrm{current\_time} > 0$}{
            \texttt{\\}
            \tcp{Sampling and mutation}
            Sample $N$ policies $\{\pi_{\theta_i}\}_{i = 1,N}$ in repertoire $\mathbb{M}$\;
            Sample batches of transitions in replay buffer $\mathbb{B}$\;
            Update half the policies using the critic $Q_v$\;
            Update the other half with genetic updates\;
            
            \texttt{\\}
            \tcp{Train the critic according to TD3 rule}
            Update the critic $Q_v$\;
            
        }
        \tcp{Evaluation and insertion in repertoire}
        Evaluate the new policies and store collected transitions in $\mathbb{B}$\;
        Update current\_time\;
        Add the updated policies in the repertoire $\mathbb{M}$\ when relevant;
        
        \texttt{\\}
        \tcp{Training the behavior extraction model}
        \If{$\textrm{current\_time in train\_times}$}{
            Update $\xi_\theta$ to reconstruct trajectories of policies in $\mathbb{M}$\;
            Re-evaluate fitness and behavior descriptor of the stored policies\;
            Update repertoire $\mathbb{M}$ with new behavior descriptors\;
        }
    }

    \caption{PGA-AURORA}
    \label{alg:PGAAURORA}
\end{algorithm}

\makeatletter
\makeatother

\SetKwComment{Comment}{/* }{ */}

\SetArgSty{textnormal}

\begin{algorithm}
    \small
    \SetAlgoLined
    \DontPrintSemicolon
    \SetKwInput{KwInput}{Given}
    \KwInput{
    \begin{itemize}
        \item Time allocation max\_time, steps per episode episode\_length
        \item Discriminator model $q_\phi$, Replay Buffer $\mathbb{B}$
        \item Diversity reward scale $\beta$
        \item Skill-conditioned policy $\pi_\theta$ and critic $Q_v$, skill prior distribution $p(z)$;
    \end{itemize}
    }
    \texttt{\\}

    \tcp{Main loop}
    $\textrm{current\_time} \leftarrow 0$\;
    \While{$\textrm{current\_time} < \textrm{max\_time}$}{
        Sample skill $z \sim p(z)$ and initial state $s_0$\;
        
        \For{$t \leftarrow 1 ... \textrm{episode\_length} $}{
            \texttt{\\}
            \tcp{Data collection in the environment}
            Sample action according to current skill $z$\;
            Take a step in the environment\;
            
            \texttt{\\}
            \tcp{Compute new rewards}
            Sample batch of transitions from buffer $\mathbb{B}$\;
            Compute the diversity rewards with the discriminator $r_{\text{diversity}} = \log q_\phi(z | s_{t+1}) - \log p(z)$\;
            Update reward by combining extrinsic and diversity rewards: $\widetilde{r_t} = r_t + \beta~r_{\text{diversity}}$\;
            
            \texttt{\\}
            \tcp{Update discriminator, policy and critic}
            Update policy parameters $\theta$ to maximize collected rewards\;
            Update discriminator parameters $\phi$ to maximize skill/state likelihood\;
            Update critic parameters $v$ to estimate state/action value\;
        }
    }

    \caption{DIAYN+Reward}
    \label{alg:DIAYN_REWARD}
\end{algorithm}

\makeatletter
\makeatother

\SetKwComment{Comment}{/* }{ */}

\SetArgSty{textnormal}

\begin{algorithm}
    \small
    \SetAlgoLined
    \DontPrintSemicolon
    \SetKwInput{KwInput}{Given}
    \KwInput{
    \begin{itemize}
        \item Time allocation max\_time, steps per episode episode\_length
        \item Discriminator model $q_\phi$, Replay Buffer $\mathbb{B}$
        \item Diversity reward scale $\beta$, reward target $R^*$ and threshold $\epsilon$
        \item Skill-conditioned policy $\pi_\theta$ and critic $Q_v$, skill prior distribution $p(z)$;
    \end{itemize}
    }
    \texttt{\\}

    \tcp{Main loop}
    $\textrm{current\_time} \leftarrow 0$\;
    \While{$\textrm{current\_time} < \textrm{max\_time}$}{
    Sample skill $z \sim p(z)$ and initial state $s_0$\;
    
    \For{$t \leftarrow 1 ... \textrm{episode\_length} $}{
            \texttt{\\}
            \tcp{Data collection in the environment}
            Sample action according to current skill $z$\;
            Take a step in the environment\;
            
            \texttt{\\}
            \tcp{Compute new rewards}
            Sample batch of transitions from buffer $\mathbb{B}$\;
            Compute the diversity rewards with the discriminator $r_{\text{diversity}} = \log q_\phi(z | s_{t+1}) - \log p(z)$\;
            Compute sum of extrinsic rewards during the episode associated to each transition $R = \sum_t r_t$\;
            Compute threshold condition $\delta = \mathbb{1}_{R > R^* - \epsilon}$\;
            Update reward by combining extrinsic and diversity rewards: $r_\text{SMERL} = r_t + \delta~\beta~r_\text{diversity}$\;
            
            \texttt{\\}
            \tcp{Update discriminator, policy and critic}
            Update policy parameters $\theta$ to maximize collected rewards\;
            Update discriminator parameters $\phi$ to maximize skill/state likelihood\;
            Update critic parameters $v$ to estimate state/action value\;
        }
    
    }

    \caption{SMERL(DIAYN)}
    \label{alg:SMERL_DIAYN}
\end{algorithm}

\makeatletter
\makeatother

\SetKwComment{Comment}{/* }{ */}

\SetArgSty{textnormal}

\begin{algorithm}
    \small
    \SetAlgoLined
    \DontPrintSemicolon
    \SetKwInput{KwInput}{Given}
    \KwInput{
    \begin{itemize}
        \item Time allocation max\_time, steps per episode episode\_length
        \item Skill-dynamics model $q_\phi$, Replay Buffer $\mathbb{B}$
        \item Diversity reward scale $\beta$
        \item Skill-conditioned policy $\pi_\theta$ and critic $Q_v$, skill prior distribution $p(z)$;
    \end{itemize}
    }
    \texttt{\\}

    \tcp{Main loop}
    $\textrm{current\_time} \leftarrow 0$\;
    \While{$\textrm{current\_time} < \textrm{max\_time}$}{
        Sample skill $z \sim p(z)$ and initial state $s_0$\;
        
        \For{$t \leftarrow 1 ... \textrm{episode\_length} $}{
            \texttt{\\}
            \tcp{Data collection in the environment}
            Sample action according to current skill $z$\;
            Take a step in the environment\;
            
            \texttt{\\}
            \tcp{Compute new rewards}
            Sample batch of transitions from buffer $\mathbb{B}$\;
            Compute the diversity rewards with the skill-dynamics $r_{\text{diversity}} = \log (q(s_{t+1}|s_t,z)) - \log (p(s))$\;
            Update reward by combining extrinsic and diversity rewards: $\widetilde{r_t} = r_t + \beta~r_{\text{diversity}}$\;
            
            \texttt{\\}
            \tcp{Update discriminator, policy and critic}
            Update policy parameters $\theta$ to maximize collected rewards\;
            Update skill-dynamics model parameters $\phi$ to maximize dynamics prediction accuracy\;
            Update critic parameters $v$ to estimate state/action value\;
        }
    
    }

    \caption{DADS+Reward}
    \label{alg:DADS_REWARD}
\end{algorithm}

\makeatletter
\makeatother

\SetKwComment{Comment}{/* }{ */}

\SetArgSty{textnormal}

\begin{algorithm}
    \small
    \SetAlgoLined
    \DontPrintSemicolon
    \SetKwInput{KwInput}{Given}
    \KwInput{
    \begin{itemize}
        \item Time allocation max\_time, steps per episode episode\_length
        \item Skill-dynamics model $q_\phi$
        \item Diversity reward scale $\beta$, reward target $R^*$ and threshold $\epsilon$
        \item Skill-conditioned policy $\pi_\theta$ and critic $Q_v$, skill prior distribution $p(z)$;
    \end{itemize}
    }
    \texttt{\\}

    \tcp{Main loop}
    $\textrm{current\_time} \leftarrow 0$\;
    \While{$\textrm{current\_time} < \textrm{max\_time}$}{
        Sample skill $z \sim p(z)$ and initial state $s_0$\;
    
        \For{$t \leftarrow 1 ... \textrm{episode\_length} $}{
            \texttt{\\}
            \tcp{Data collection in the environment}
            Sample action according to current skill $z$\;
            Take a step in the environment\;
            
            \texttt{\\}
            \tcp{Compute new rewards}
            Sample batch of transitions from buffer $\mathbb{B}$\;
            Compute the diversity rewards with the skill-dynamics $r_{\text{diversity}} = \log (q(s_{t+1}|s_t,z)) - \log (p(s))$\;
            Compute sum of extrinsic rewards during the episode associated to each transition $R = \sum_t r_t$\;
            Compute threshold condition $\delta = \mathbb{1}_{R > R^* - \epsilon}$\;
            Update reward by combining extrinsic and diversity rewards: $r_\text{SMERL} = r_t + \delta~\beta~r_\text{diversity}$\;
            
            \texttt{\\}
            \tcp{Update discriminator, policy and critic}
            Update policy parameters $\theta$ to maximize collected rewards\;
            Update skill-dynamics model parameters $\phi$ to maximize dynamics prediction accuracy\;
            Update critic parameters $v$ to estimate state/action value\;
        }
    
    }

    \caption{SMERL(DADS)}
    \label{alg:SMERL_DADS}
\end{algorithm}

\end{document}